\documentclass[10pt,twocolumn,letterpaper]{article}

\usepackage{wacv}
\usepackage{times}
\usepackage{epsfig}
\usepackage{graphicx}
\usepackage{amsmath}
\usepackage{amssymb}

\usepackage{dsfont}
\usepackage{subcaption}
\usepackage{mathtools}
\usepackage[]{algorithm2e}
\usepackage{flushend}
\usepackage[numbers,sort]{natbib}
\usepackage{booktabs}
\usepackage[table]{xcolor}
\usepackage{multicol}
\usepackage{enumerate}

\definecolor{lightgray}{gray}{0.95}

\usepackage[pagebackref=true,breaklinks=true,letterpaper=true,colorlinks,bookmarks=false]{hyperref}

\wacvfinalcopy 

\def\wacvPaperID{724} 

\ifwacvfinal\fi
\begin{document}
	
	\title{Probabilistic Object Detection: Definition and Evaluation}
	
	\author{
		David Hall$^{1,2}$ \space Feras Dayoub$^{1,2}$ \space John Skinner$^{1,2}$ \space Haoyang Zhang$^{1,2}$ \space Dimity Miller$^{1,2}$ \\ \space Peter Corke$^{1,2}$ 
		Gustavo Carneiro$^{1,3}$ \space Anelia Angelova$^{4}$ \space Niko S\"{u}nderhauf$^{1,2}$\\
		$^1$Australian Centre for Robotic Vision\\
		$^2$Queensland University of Technology \space $^3$University of Adelaide \space $^4$Google Brain\\
		$^2${\tt\small\{d20.hall, feras.dayoub, j6.skinner, haoyang.zhang.acrv,} \\{\ttfamily\small d24.miller, peter.corke, niko.suenderhauf\}@qut.edu.au}\\
		$^3${\tt\small gustavo.carneiro@adelaide.edu.au} \space
		$^4${\tt\small anelia@google.com}\\
	}
	
	\maketitle
	\ifwacvfinal\thispagestyle{empty}\fi
	
	\begin{abstract}
		We introduce Probabilistic Object Detection, the task of detecting objects in images and accurately quantifying the spatial and semantic uncertainties of the detections. 
		Given the lack of methods capable of assessing such probabilistic object detections, we present the new Probability-based Detection Quality measure (PDQ).
		Unlike AP-based measures, PDQ has no arbitrary thresholds and rewards spatial and label quality, and foreground/background separation quality while explicitly penalising false positive and false negative detections. 
		We contrast PDQ with existing mAP and moLRP measures by evaluating state-of-the-art detectors and a Bayesian object detector based on Monte Carlo Dropout. Our experiments indicate that conventional object detectors tend to be spatially overconfident and thus perform poorly on the task of probabilistic object detection. 
		Our paper aims to encourage the development of new object detection approaches that provide detections with accurately estimated spatial and label uncertainties and are of critical importance for deployment on robots and embodied AI systems in the real world.
	\end{abstract}

\newcommand{\vect}[1]{\mathbf{ #1}}

\newcommand{\vectg}[1]{{\boldsymbol{ #1}}}

\newcommand{\ggo}{\ensuremath{\mathrm{g^2o}} }

\newcommand{\R}{\mathbb{R}}
\newcommand{\N}{\mathbb{N}}
\newcommand{\Z}{\mathbb{Z}}
\renewcommand{\P}{\mathbb{P}}

\newcommand{\tran}{^\top}

\newcommand{\T}{^\mathsf{T}}
\newcommand{\iT}{^{-\mathsf{T}}}

\newcommand{\inv}{^{-1}}

\newcommand{\func}[2]{\mathtt{#1}\left\{#2\right\}}

\newcommand{\sig}{\operatorname{sig}}

\newcommand{\diag}{\operatorname{diag}}

\newcommand{\argmin}{\operatornamewithlimits{argmin}}

\newcommand{\argmax}{\operatornamewithlimits{argmax}}

\newcommand{\RMSE}{\operatorname{RMSE}}
\newcommand{\RMSEpos}{\operatorname{RMSE}_\text{pos}}
\newcommand{\RMSEori}{\operatorname{RMSE}_\text{ori}}

\newcommand{\RPE}{\operatorname{RPE}}
\newcommand{\RPEpos}{\operatorname{RPE}_\text{pos}}
\newcommand{\RPEori}{\operatorname{RPE}_\text{ori}}

\newcommand{\rpe}{\varepsilon_{\vdelta}}

\newcommand{\achiError}{\bar{e}_{\chi^2}}

\newcommand{\chiError}{e_{\chi^2}}

\newcommand{\normal}[2]{\mathcal{N}\left(#1, #2\right)}

\newcommand{\uniform}[2]{\mathcal{U}\left(#1, #2\right)}

\newcommand{\pfrac}[2]{\frac{\partial #1}{\partial #2}}  
\newcommand{\fracpd}[2]{\frac{\partial #1}{\partial #2}} 
\newcommand{\fracppd}[2]{\frac{\partial^2 #1}{\partial #2^2}}  

\newcommand{\dd}{\mathrm{d}}  

\newcommand{\smd}[2]{\left\| #1 \right\|^2_{#2}}

\newcommand{\E}[1]{\text{\normalfont{E}}\left[ #1 \right]}     
\newcommand{\Cov}[1]{\text{\normalfont{Cov}}\left[ #1 \right]} 
\newcommand{\Var}[1]{\text{\normalfont{Var}}\left[ #1 \right]} 
\newcommand{\Tr}[1]{\text{\normalfont{tr}}\left( #1 \right)}   
\def\sgn{\mathop{\mathrm sgn}}

\newcommand{\twovector}[2]{\begin{pmatrix} #1 \\ #2 \end{pmatrix}} 
\newcommand{\smalltwovector}[2]{\left(\begin{smallmatrix} #1 \\ #2 \end{smallmatrix}\right)} 
\newcommand{\threevector}[3]{\begin{pmatrix} #1 \\ #2 \\ #3 \end{pmatrix}} 
\newcommand{\fourvector}[4]{\begin{pmatrix} #1 \\ #2 \\ #3 \\ #4 \end{pmatrix}}  

\newcommand{\smallthreevector}[3]{\left(\begin{smallmatrix} #1 \\ #2 \\ #3 \end{smallmatrix}\right)} 

\newcommand{\fourmatrix}[4]{\begin{pmatrix} #1 & #2 \\ #3 & #4 \end{pmatrix}} 

\newcommand{\vA}{\vect{A}}
\newcommand{\vB}{\vect{B}}
\newcommand{\vC}{\vect{C}}
\newcommand{\vD}{\vect{D}}
\newcommand{\vE}{\vect{E}}
\newcommand{\vF}{\vect{F}}
\newcommand{\vG}{\vect{G}}
\newcommand{\vH}{\vect{H}}
\newcommand{\vI}{\vect{I}}
\newcommand{\vJ}{\vect{J}}
\newcommand{\vK}{\vect{K}}
\newcommand{\vL}{\vect{L}}
\newcommand{\vM}{\vect{M}}
\newcommand{\vN}{\vect{N}}
\newcommand{\vO}{\vect{O}}
\newcommand{\vP}{\vect{P}}
\newcommand{\vQ}{\vect{Q}}
\newcommand{\vR}{\vect{R}}
\newcommand{\vS}{\vect{S}}
\newcommand{\vT}{\vect{T}}
\newcommand{\vU}{\vect{U}}
\newcommand{\vV}{\vect{V}}
\newcommand{\vW}{\vect{W}}
\newcommand{\vX}{\vect{X}}
\newcommand{\vY}{\vect{Y}}
\newcommand{\vZ}{\vect{Z}}

\newcommand{\va}{\vect{a}}
\newcommand{\vb}{\vect{b}}
\newcommand{\vc}{\vect{c}}
\newcommand{\vd}{\vect{d}}
\newcommand{\ve}{\vect{e}}
\newcommand{\vf}{\vect{f}}
\newcommand{\vg}{\vect{g}}
\newcommand{\vh}{\vect{h}}
\newcommand{\vi}{\vect{i}}
\newcommand{\vj}{\vect{j}}
\newcommand{\vk}{\vect{k}}
\newcommand{\vl}{\vect{l}}
\newcommand{\vm}{\vect{m}}
\newcommand{\vn}{\vect{n}}
\newcommand{\vo}{\vect{o}}
\newcommand{\vp}{\vect{p}}
\newcommand{\vq}{\vect{q}}
\newcommand{\vr}{\vect{r}}
\newcommand{\vt}{\vect{t}}
\newcommand{\vu}{\vect{u}}
\newcommand{\vv}{\vect{v}}
\newcommand{\vw}{\vect{w}}
\newcommand{\vx}{\vect{x}}
\newcommand{\vy}{\vect{y}}
\newcommand{\vz}{\vect{z}}

\newcommand{\valpha}{\vectg{\alpha}}
\newcommand{\vbeta}{\vectg{\beta}}
\newcommand{\vgamma}{\vectg{\gamma}}
\newcommand{\vdelta}{\vectg{\delta}}
\newcommand{\vepsilon}{\vectg{\epsilon}}
\newcommand{\vtau}{\vectg{\tau}}
\newcommand{\vmu}{\vectg{\mu}}
\newcommand{\vphi}{\vectg{\phi}}
\newcommand{\vPhi}{\vectg{\Phi}}
\newcommand{\vpi}{\vectg{\pi}}
\newcommand{\vPi}{\vectg{\Pi}}
\newcommand{\vPsi}{\vectg{\Psi}}
\newcommand{\vchi}{\vectg{\chi}}
\newcommand{\vvarphi}{\vectg{\varphi}}
\newcommand{\veta}{\vectg{\eta}}
\newcommand{\viota}{\vectg{\iota}}
\newcommand{\vkappa}{\vectg{\kappa}}
\newcommand{\vlambda}{\vectg{\lambda}}
\newcommand{\vLambda}{\vectg{\Lambda}}
\newcommand{\vnu}{\vectg{\nu}}
\newcommand{\vgo}{\vectg{\o}}
\newcommand{\vvarpi}{\vectg{\varpi}}
\newcommand{\vtheta}{\vectg{\theta}}
\newcommand{\vvartheta}{\vectg{\vartheta}}
\newcommand{\vrho}{\vectg{\rho}}
\newcommand{\vsigma}{\vectg{\sigma}}
\newcommand{\vSigma}{\vectg{\Sigma}}
\newcommand{\vvarsigma}{\vectg{\varsigma}}
\newcommand{\vupsilon}{\vectg{\upsilon}}
\newcommand{\vomega}{\vectg{\omega}}
\newcommand{\vOmega}{\vectg{\Omega}}
\newcommand{\vxi}{\vectg{\xi}}
\newcommand{\vXi}{\vectg{\Xi}}
\newcommand{\vpsi}{\vectg{\psi}}
\newcommand{\vzeta}{\vectg{\zeta}}

\newcommand{\vzero}{\vect{0}}

\newcommand{\cA}{\mathcal{A}}
\newcommand{\cB}{\mathcal{B}}
\newcommand{\cC}{\mathcal{C}}
\newcommand{\cD}{\mathcal{D}}
\newcommand{\cE}{\mathcal{E}}
\newcommand{\cF}{\mathcal{F}}
\newcommand{\cG}{\mathcal{G}}
\newcommand{\cH}{\mathcal{H}}
\newcommand{\cI}{\mathcal{I}}
\newcommand{\cJ}{\mathcal{J}}
\newcommand{\cK}{\mathcal{K}}
\newcommand{\cL}{\mathcal{L}}
\newcommand{\cM}{\mathcal{M}}
\newcommand{\cN}{\mathcal{N}}
\newcommand{\cO}{\mathcal{O}}
\newcommand{\cP}{\mathcal{P}}
\newcommand{\cQ}{\mathcal{Q}}
\newcommand{\cR}{\mathcal{R}}
\newcommand{\cS}{\mathcal{S}}
\newcommand{\cT}{\mathcal{T}}
\newcommand{\cU}{\mathcal{U}}
\newcommand{\cV}{\mathcal{V}}
\newcommand{\cW}{\mathcal{W}}
\newcommand{\cX}{\mathcal{X}}
\newcommand{\cY}{\mathcal{Y}}
\newcommand{\cZ}{\mathcal{Z}}

\newcommand{\fA}{\mathfrak{A}}
\newcommand{\fB}{\mathfrak{B}}
\newcommand{\fC}{\mathfrak{C}}
\newcommand{\fD}{\mathfrak{D}}
\newcommand{\fE}{\mathfrak{E}}
\newcommand{\fF}{\mathfrak{F}}
\newcommand{\fG}{\mathfrak{G}}
\newcommand{\fH}{\mathfrak{H}}
\newcommand{\fI}{\mathfrak{I}}
\newcommand{\fJ}{\mathfrak{J}}
\newcommand{\fK}{\mathfrak{K}}
\newcommand{\fL}{\mathfrak{L}}
\newcommand{\fM}{\mathfrak{M}}
\newcommand{\fN}{\mathfrak{N}}
\newcommand{\fO}{\mathfrak{O}}
\newcommand{\fP}{\mathfrak{P}}
\newcommand{\fQ}{\mathfrak{Q}}
\newcommand{\fR}{\mathfrak{R}}
\newcommand{\fS}{\mathfrak{S}}
\newcommand{\fT}{\mathfrak{T}}
\newcommand{\fU}{\mathfrak{U}}
\newcommand{\fV}{\mathfrak{V}}
\newcommand{\fW}{\mathfrak{W}}
\newcommand{\fX}{\mathfrak{X}}
\newcommand{\fY}{\mathfrak{Y}}
\newcommand{\fZ}{\mathfrak{Z}}

	\section{Introduction}\label{sec:intro}
Visual object detection provides answers to two questions: \emph{what} is in an image and \emph{where} is it? State-of-the-art approaches that address this problem are based on deep convolutional neural networks (CNNs) that localise objects by predicting a bounding box, and providing a class label with a confidence score, or a full label distribution, for every detected object in the image~\cite{redmon_yolov3:_2018,ren2015faster,liu2016ssd}.
The ability of deep CNNs to quantify epistemic and aleatoric uncertainty~\cite{kendall2017uncertainties} has recently been identified as paramount for deployment in safety critical applications, where the perception and decision making of an agent has to be trusted~\cite{suenderhauf2018limits,kendall2017uncertainties, amodei2016concrete, varshney17on}. 
While state-of-the-art object detectors have limited capability to express epistemic and aleatoric uncertainty about the class label through the confidence score or label distribution ~\cite{Guo2017OnCO,nguyen2015deep, hendrycks17baseline, szegedy2014intriguing, torralba11unbiased}, uncertainty about the spatial aspects of the detection is currently not at all quantified. 
\begin{figure}[t]
    \centering
    \includegraphics[width=0.8\linewidth]{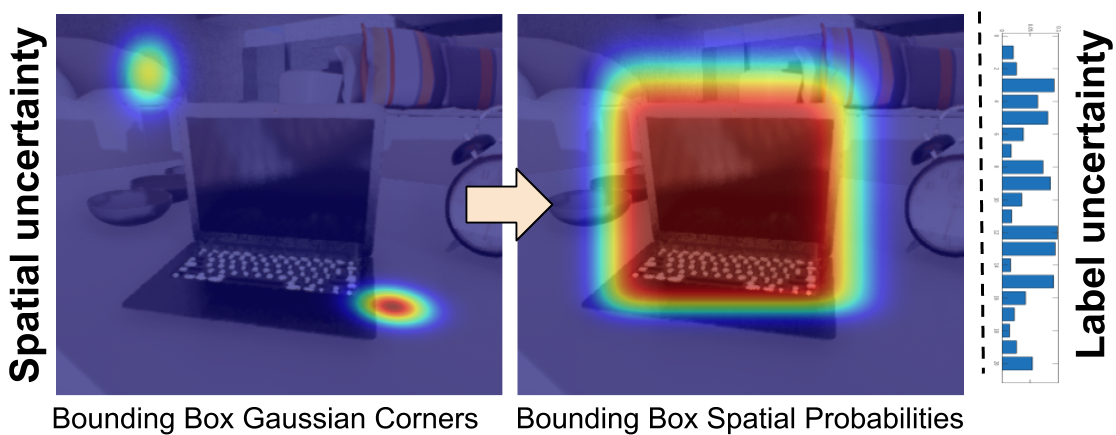}
    \caption{
    In contrast to conventional object detection, \textit{probabilistic} object detections express semantic \textit{and} spatial uncertainty.
    Our probabilistic object detections represent object locations as probabilistic bounding boxes where corners are modelled as 2D Gaussians (left) used to express a spatial uncertainty over the pixels (centre).
    Semantic uncertainty is represented as full label probability distributions (right). 
    }
    \vspace{-8pt}
    \label{fig:corners}
\end{figure}
Furthermore, none of the existing benchmarks using average precision (AP) as the basis for their evaluation~\cite{everingham_pascal_2010, everingham2015pascal,russakovsky_imagenet_2015,lin_microsoft_2014,OpenImages,OpenImages2} can evaluate how well detectors quantify spatial and semantic uncertainties.

We introduce \textbf{Probabilistic Object Detection}, the task of detecting objects in images while accurately quantifying the spatial and semantic uncertainties of the detections. Probabilistic Object Detection poses a key challenges that goes beyond the established conventional object detection: the detector must quantify its \emph{spatial uncertainty} by reporting \emph{probabilistic} bounding boxes, where the box corners are modelled as normally distributed. As illustrated in Figure~\ref{fig:corners}, this induces a spatial probability distribution over the image for each detection.
The detector must also reliably quantify its \emph{semantic uncertainty} by providing a full probability distribution over the known classes for each detection. 

To evaluate how well detectors perform on this challenging task, we introduce a new evaluation measure, \textbf{Probability-based Detection Quality} (PDQ). 
In contrast to AP-based measures, PDQ explicitly evaluates the reported probability of the true class via its \emph{Label Quality} component. 
Furthermore, PDQ contains a \emph{Spatial Quality} term that evaluates how well a detection's spatial probability distribution matches the true object.

Unlike existing measures such as mAP~\cite{lin_microsoft_2014} and moLRP~\cite{oksuz2018localization}, PDQ jointly evaluates
spatial and label uncertainty quality, foreground and background separation quality, and the number of true positive (correct), false positive (spurious), and false negative (missed) detections. Importantly, PDQ does not rely on fixed thresholds or tuneable parameters, and provides optimal assignments of detections to ground-truth objects.
Although PDQ has been primarily developed to evaluate \emph{new} types of probabilistic object detectors that are designed to quantify spatial and semantic uncertainties, PDQ can also evaluate conventional state-of-the-art, non-probabilistic detectors. 

As we show in Section \ref{sec:eval_real_detectors}, current conventional detection methods perform poorly on the task of probabilistic object detection due to spatial over-confidence and are outperformed by a recently proposed probabilistic object detector that incorporates Monte Carlo Dropout into a VGG16-based Single Shot MultiBox Detector (SSD)~\cite{miller2018evaluating}.

In summary, our contributions include defining the challenging new task of probabilistic object detection, introducing the new evaluation measure PDQ, evaluating current object detectors, and showing for the first time that novel probabilistic object detectors achieve better performance on this new task, that is highly relevant for applications such as robotics or embodied AI.

\section{Motivation}
\label{sec:motivation}
Object detection embedded in a robot or autonomous system, such as a self-driving car, is part of a complex, active, goal-driven system. In such a scenario, object detection provides crucial perception information that ultimately determines the performance of the robot in its environment. Mistakes in object detection could lead to catastrophic outcomes that not only risk the success of the robot's mission, but potentially endanger human lives~\cite{namba2018risks, amodei2016concrete, otte13safe, varshney17on,cirecsan2012multi,lenz2015deep}. 

For safe and trusted operation in robots or autonomous systems, CNNs must express meaningful \emph{uncertainty} information~\cite{suenderhauf2018limits, kendall2017uncertainties, hendrycks17baseline, Guo2017OnCO, amodei2016concrete, varshney17on}. Object detectors will have to quantifying uncertainty for both the reported labels and bounding boxes, which would enable them to be treated as yet another sensor within the established and trusted framework of Bayesian information fusion~\cite{thrun2005probabilistic,richter2018bayesian}. However, while state-of-the-art object detectors report at least an \emph{uncalibrated} indicator of label uncertainty via label distributions or label scores~\cite{Guo2017OnCO,nguyen2015deep, hendrycks17baseline, szegedy2014intriguing, torralba11unbiased}, they currently do \emph{not} report spatial uncertainty. As a result, evaluating the quality of the label or spatial uncertainties is not within the scope of typical benchmark measures and competitions~\cite{everingham_pascal_2010, everingham2015pascal,russakovsky_imagenet_2015,lin_microsoft_2014,OpenImages,OpenImages2}.

We argue in favour of accurate quantification of spatial and semantic uncertainties for object detectors in computer vision and robotics applications. Our work builds on this idea by creating a measure that will guide research towards developing detection systems that can operate effectively within a robot's sensor fusion framework.

	\section{Related Work}\label{sec:lit}

\paragraph{Conventional Object Detection:}
Object detection is a fundamental task in computer vision and aims to localise each instance of certain object classes in an image using a bounding box. The typical output from an object detection system is a set of bounding boxes with a class label score~\cite{viola2004robust, dalal2005histograms, DPM}. Since the advent of convolutional neural networks (CNNs)~\cite{krizhevsky_imagenet_2012}, object detection has experienced impressive progress in terms of accuracy and speed~\cite{RCNN, girshick2015fast, ren2015faster,liu2016ssd, redmon_yolov3:_2018, dai2016r, lin2018focal}. Nonetheless, current overconfident object detection systems fail to provide spatial and semantic uncertainties, and as a result, can be a source of risk in various vision and robotics applications.
The probabilistic object detection task introduced by this paper requires that object detectors estimate the spatial and semantic uncertainty of their detections.

\vspace{-1.5em}
\paragraph{Uncertainty Estimation:}
To improve system robustness and accuracy or avoid risks, quantifying uncertainty has become popular in many vision tasks. 
Kendall et al.~\cite{kendall2015bayesian} propose a Bayesian model that outputs a pixel-wise semantic segmentation with a measure of model uncertainty for each class. In~\cite{kendall2017uncertainties}, the authors propose to model the aleatoric and epistemic uncertainties for the pixel-wise semantic segmentation and depth regression tasks, and argue that epistemic uncertainty is important for safety-critical applications and training with small data sets. Kampffmeyer et al.~\cite{kampffmeyer2016semantic} propose a model that estimates pixel-wise classification uncertainty in urban remote sensing images -- they argue that the estimated uncertainty can indicate the correctness of pixel labelling. Miller et al.~\cite{miller2018dropout, miller2018evaluating} estimate both spatial and classification uncertainties for object detection and use the uncertainty to accept or reject detections under open-set conditions. Nair et al.~\cite{nair2018exploring} provide four different voxel-based uncertainty measures for their 3D lesion segmentation system to enable a more complete revision by clinicians. In~\cite{tanno2017bayesian} an uncertainty map for super-resolution of diffusion MR brain images is generated  to enable a risk assessment for the clinical use of the super-resolved images. In~\cite{Nuclei}, the authors build an ensemble of predictors to estimate the uncertainty of the centre of nuclei in order to produce more accurate classification results. 
All the methods above, except the last one~\cite{Nuclei}, estimate uncertainty based on the Monte Carlo (MC) dropout technique~\cite{gal2015bayesian, gal2016dropout}. 
The papers above provide evidence that it is important to estimate  uncertainty for various vision tasks. Most of the proposed methods, except~\cite{miller2018evaluating, Nuclei}, deal with pixel-wise classification.We argue that it is essential to capture the uncertainty of object detectors as motivated in Section~\ref{sec:motivation}.
\vspace{-1.5em}
\paragraph{Performance Measures:}
For the past decade, detection algorithms have predominantly been evaluated using average precision (AP) or variants thereof.
Average precision was introduced for the PASCAL VOC challenge~\cite{everingham_pascal_2010} in 2007 to replace measuring the area under the ROC curve.
It is the average of the maximum precision values at different recall values.
These use a pre-defined threshold for the intersection over union (IoU), typically 0.5, defining a true positive detection.
This is calculated and averaged across all classes.
Since then, AP has become the standard evaluation measure in the PASCAL VOC challenge and is the basis for many other works examining object detection~\cite{lin_microsoft_2014, russakovsky_imagenet_2015, liu2016ssd, lin2018focal, ren2015faster, dai2016r}. 
Most recently, a variation of AP was created which averages AP over multiple IoU thresholds (varying from 0.5 to 0.95 in intervals of 0.05)~\cite{lin_microsoft_2014}.
This averaging over IoUs rewards detectors with better localisation accuracy.
In this work we refer to this measure as mean average precision (mAP) to distinguish it from AP despite mAP typically referring to averaging AP over all classes.

AP-based measures have biased the community to develop object detectors with high recall rate and localisation precision, but these measures have several weaknesses~\cite{hoiem2012diagnosing, csurka2013good, pont2016supervised}. They rely on fixed IoU thresholds which can lead to overfitting for certain IoU thresholds -- the negative consequence is that a small change in the thresholds can cause abrupt score changes. Additionally, these measures use the label score as the detection ranking evidence, without considering the spatial quality, which can lead to sub-optimal detection assignment.
In our work, we propose the new evaluation measure PDQ to evaluate both label and spatial qualities of object detections, without using any fixed thresholds and relying on an optimal assignment of detection to ground-truth objects based on both spatial and label qualities.

Oksuze et al.~\cite{oksuz2018localization} propose the Localisation Recall Precision (LRP) metric to overcome two main deficiencies of mAP: the inability to distinguish different precision-recall (PR) curves, and the lack of a direct way to measure bounding box localisation accuracy. 
When used for analysing multi-class detectors, the mean optimal LRP (moLRP) is used.
Comparing to mAP, moLRP is also based on PR curves but measures localisation quality, false positive rate and false negative rate at some optimal label threshold for each class.
The localisation quality is represented by the IoU between the detection and the ground-truth object, scaled by the IoU threshold being used to plot the PR curves. In contrast, our PDQ measure 
estimates the spatial uncertainty through probabilistic bounding boxes and evaluates how well the detection bounding box's spatial probability distribution coincides with the true object.

\section{Probabilistic Object Detection}
\label{sec:probabilistic_object_detection}

Probabilistic Object Detection is the task of detecting objects in an image, while accurately quantifying the spatial and semantic uncertainties of the detections. Probabilistic Object Detection thus extends conventional object detection, and makes the quantification of uncertainty an essential part of the task and its evaluation.

Probabilistic Object Detection requires a detector to provide for each known object in an image:
\begin{itemize}
    \item \vspace{-0.5em} a categorical distribution over all class labels, and
    \item \vspace{-0.5em} a bounding box represented as $\cB = (\cN_0, \cN_1) = \left(\normal{\vmu_0}{\vSigma_0}, \normal{\vmu_1}{\vSigma_1}\right)$ such that $\vmu_i$ and $\vSigma_i$ are the mean and covariances for the multivariate Gaussians describing the top-left and bottom-right corner of the box.
\end{itemize}
\vspace{-0.5em} \noindent From this probabilistic box representation $\cB$, we can calculate a probability distribution $P$ over all pixels $(u',v')$, such that $P(u',v')$ is the probability that the pixel is contained in the box:
\begin{align*}
\begin{split}
    P(u',v') = \iint_{0,0}^{v',u'} \cN_0(u, v) \,du\,dv  \iint_{v', u'}^{H,W} \cN_1(u, v) \,du\,dv,
\end{split}
\end{align*}
where $H,W$ is the height and width of the image.
This is illustrated in Fig. \ref{fig:corners}, with Gaussians over two corners illustrated on the left, and the resulting distribution $P(u',v')$ in the centre.

The evaluation of each detection focuses on the probability value assigned to the true class label, and the spatial probability mass from $P(u',v')$ assigned to the ground truth object \vs the probability mass assigned to the background.  Since existing measures for conventional object detection such as mAP~\cite{lin_microsoft_2014} or moLRP~\cite{oksuz2018localization} are not equipped to evaluate the probabilistic aspects of a detection, we introduce a novel evaluation measure for Probabilistic Object Detection in the following section.

\section{Probability-based Detection Quality (PDQ)}\label{sec:metric_def}
This section introduces the major technical contribution of our paper: the probability-based detection quality (PDQ) measure which evaluates the quality of detections based on spatial and label probabilities.
Unlike AP-based measures, our approach penalises low spatial uncertainty when detecting background as foreground, or when detecting foreground as background, and explicitly evaluates the label probability in calculating detection quality. 
PDQ has no thresholds or tuneable parameters that can redefine the conditions of success. 
Furthermore, PDQ is based on an approach that provides optimal assignment of detections to ground-truth objects, incorporating both the label and spatial attributes of the detections in this assignment.

A reference implementation of PDQ will be made available on github (link withheld for double-blind review).
\vspace{-1em}
\paragraph{Notation}\label{sec:def:gto_det_defs}
We write the $i$-th ground-truth object in the $f$-th frame (image) as the set $\cG^f_i = \{ \hat\cS^f_i, \hat\cB^f_i, \hat{c}^f_i \}$, comprising a segmentation mask defined by a set of pixels $\hat{\cS}^f_i$, a set of bounding box corners $\hat{\cB}^f_i$ fully encapsulating all pixels in $\hat{\cS}^f_i$, and a class label $\hat{c}^f_i$.

\begin{figure}[t]
\newlength{\twosubht}
\newsavebox{\twosubbox}
\sbox\twosubbox{%
  \resizebox{\dimexpr0.95\linewidth-1em}{!}{%
    \includegraphics[height=3cm]{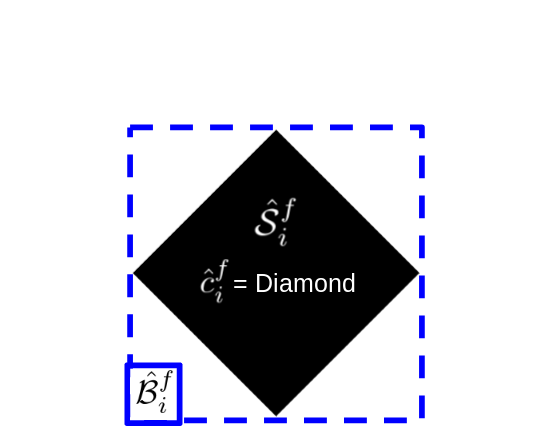}%
    \includegraphics[height=3cm]{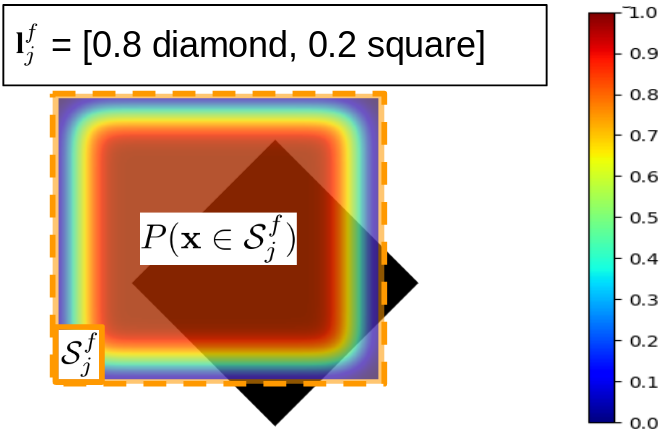}%
  }%
}
\setlength{\twosubht}{\ht\twosubbox}

\centering

\subcaptionbox{ground-truth object\label{fig:ground_truth_object}}{%
  \includegraphics[height=\twosubht]{imgs/gt_def_img.png}%
}\quad
\subcaptionbox{detection\label{fig:detection}}{%
  \includegraphics[height=\twosubht]{imgs/det_def_img.png}%
}
\caption{
In our notation, a ground-truth object (a) consists of a segmentation mask $\hat{\cS}^f_i$ (black), a bounding box $\hat{\cB}^f_i$ (blue box), and a class label $\hat{c}^f_i$ which here is \emph{diamond}. A detection (b) consists of a probability density function $P(\vx \in \cS^f_j)$ (illustrated as a heatmap), a segmentation mask $\cS^f_j$ (all pixels within the orange box), and a probability distribution across all classes $\textbf{l}^f_j$, which here provides probabilities for diamond and square classes.}\label{fig:example_ground_truth_and_detection}
\end{figure}

We define the $j$-th detection in the $f$-th frame as the set $\cD^f_j = \{P(\vx \in \cS^f_j), \cS^f_j , \textbf{l}^f_j\}$, comprising a probability function that returns the spatial probability that a given pixel is a part of the detection (regardless of class prediction) $P(\vx \in \cS^f_j)$, a set of pixels with a non-zero $P(\vx \in \cS^f_j)$ which we refer to as the detection segmentation mask $\cS^f_j$, and a label probability distribution across all possible class labels $\textbf{l}^f_j$. 
A visualisation of both ground-truth objects and detections is provided in Figure~\ref{fig:example_ground_truth_and_detection}.

\vspace{-1em}
\paragraph{Requirements}
PDQ requires pixel-accurate ground-truth annotations for the segmentation mask $\hat{\cS}^f_i$. Such annotations can be easily obtained from simulated environments~\cite{carla:Dosovitskiy17, minos:savva2017minos} and also from datasets containing only bounding box annotations by considering all pixels within a box part of the segmentation mask. PDQ can evaluate probabilistic detectors that provide bounding boxes with Gaussian corners as defined in Section \ref{sec:probabilistic_object_detection}, or conventional detectors by assuming $P(\vx \in \cS^f_j) = 1-\epsilon$ for all pixels inside the respective bounding box and $\epsilon$ outside, for a small $\epsilon > 0$.

\label{sec:def:pdq}
\vspace{-1em}
\paragraph{Overview}
PDQ evaluates both the \emph{spatial} and \emph{label} quality of a detector. It is therefore based on a combination of a spatial quality measure $Q_S$ and a label quality measure $Q_L$. Both are calculated between all possible pairs of detections and ground-truth objects within a single frame. We define the geometric mean between these two quality measures as the pairwise PDQ (pPDQ), and use it to find the optimal assignment between all detections and ground-truth objects within an image. The optimal pPDQ measures are then combined into an overall PDQ measure for the whole dataset. However, many of these intermediate results can also be recorded and analysed for a more detailed breakdown of performance. Algorithm~\ref{alg:pdq} summarises the overall PDQ calculation. In the following, we detail each of the involved steps and both quality measures.

\begin{algorithm}[t]
\SetAlgoLined\SetArgSty{}
\KwData{a dataset of $f=1\dots N_F$ frames with detections $\cD^f_j$ and ground-truths $\cG^f_i$}
\ForAll{frames in the dataset}{
   \ForAll{pairs $(\cG^f_i, \cD^f_j)$}{
    calculate spatial quality $Q_{S}(\cG^f_i, \cD^f_j)$
    
    calculate label quality $Q_{L}(\cG^f_i, \cD^f_j)$
    
    calculate $\operatorname{pPDQ}(\cG^f_i, \cD^f_j) = \sqrt{Q_{S} \cdot Q_{L}}$
    }
    
    Based on the $\operatorname{pPDQ}(.)$ computed between all pairs, find optimal assignment between detections and ground-truth objects, yielding optimal pPDQ for frame $f$.
}
Combine frame-wise optimal pPDQs into an overall PDQ measure.
\caption{PDQ Evaluation Process}
\label{alg:pdq}
\end{algorithm}

\subsection{Spatial Quality}
The spatial quality $Q_S$ measures how well a detection $\cD^f_j$ captures the spatial extent of a ground-truth object $\cG^f_i$, and takes into account the spatial probabilities for individual pixels as expressed by the detector. 

Spatial quality $Q_S$ comprises two loss terms, the foreground loss $L_{FG}$ and the background loss $L_{BG}$. Spatial quality is defined as the exponentiated negative sum of the two loss terms, as follows: 
\begin{equation}
    Q_S(\cG^f_i, \cD^f_j) = \exp(-(L_{FG}(\cG^f_i, \cD^f_j) + L_{BG}(\cG^f_i, \cD^f_j)),
    \label{eq:Q_S}
\end{equation}
where $Q_S(\cG^f_i, \cD^f_j) \in [0,1]$.  The spatial quality in \eqref{eq:Q_S} is equal to 1 if the detector assigns a spatial probability of 1 to all ground-truth pixels, while not assigning any probability mass to pixels outside the ground-truth segment. This behaviour is governed by the two loss terms explained below.

\vspace{-1em}
\paragraph{Foreground Loss}
The foreground loss $L_{FG}$ is defined as the average negative log-probability the detector assigns to the pixels of a ground-truth segment.
\begin{equation}
    L_{FG}(\cG^f_i, \cD^f_j) = - \frac{1}{|\hat{\cS}^f_i|} \sum_{\vx \in \hat{\cS}^f_i} \log(P(\vx \in \cS^f_j)),
    \label{eq:fgloss}
\end{equation}
where, as defined above, $\hat{\cS}^f_i$ is the set of all pixels belonging to the $i$-th ground-truth segment in frame $f$, and $P(\cdot)$ is the spatial probability function that assigns a probability value to every pixel of the $j$-th detection.
The foreground loss is minimised if the detector assigns a probability value of one to every pixel of the ground-truth segment, in which case $L_{FG}=0$. It grows without bounds otherwise. 

\begin{figure}[t]
    \centering
    \includegraphics[width=0.7\linewidth]{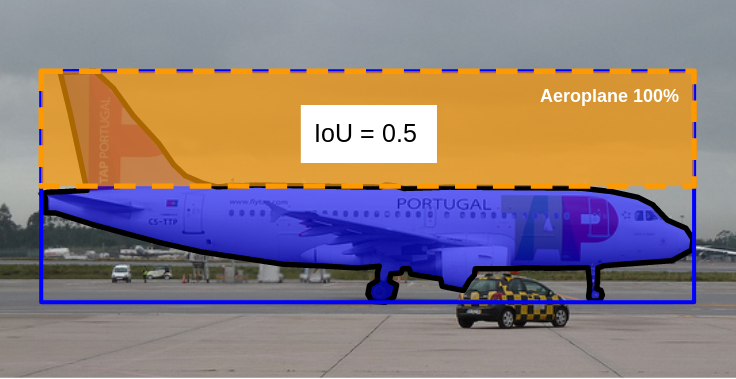}
    \caption{Example of a detection of an aeroplane (orange box), a ground-truth box (blue line), and a ground-truth segmentation mask, (blue-coloured region with black border). At an IoU threshold of 0.5, AP-based methods consider the orange detection entirely correct, despite covering only 16\% of the plane's pixels. 
    There is no correlation between the bounding box overlap analysed and the content within the bounding box. 
    By comparison, PDQ penalises this detection heavily for only detecting this small portion without any spatial uncertainty. The pPDQ for this detection containing no spatial uncertainty is $3.64 \times 10^{-6}$.}
    \label{fig:plane_iou_example}
\end{figure}

Notice that $L_{FG}$ intentionally ignores pixels that are inside the ground-truth bounding box $\hat{\cB}^f_i$ but are \emph{not} part of the ground-truth segment $\hat{\cS}^f_i$. 
This avoids treating the detection of background pixels as critically important in the case of irregularly shaped objects when pixel-level annotations are available, unlike AP-based methods using bounding-box IoUs, as illustrated in Figure~\ref{fig:plane_iou_example}.

\begin{figure}[t]
    \centering
    \includegraphics[width=0.9\linewidth]{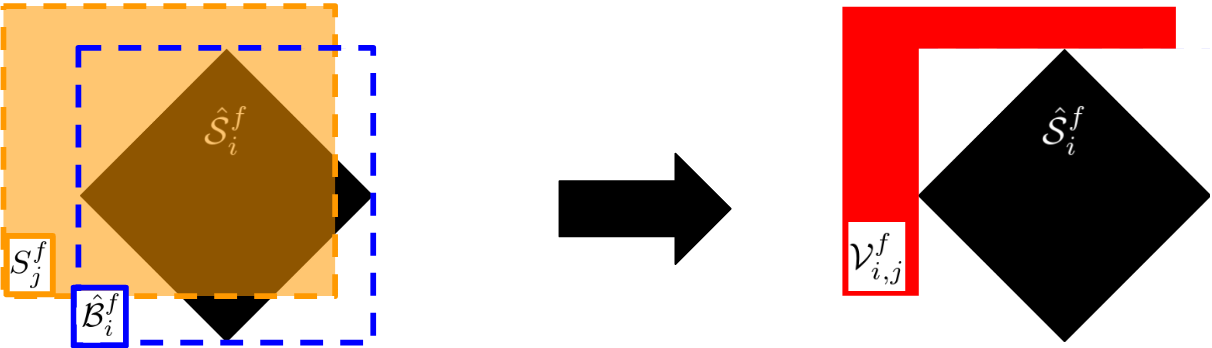}
    \caption{PDQ defines the background evaluation region $\cV^f_{i,j}$ (red) as the set of pixels that are part of the detection $\cS^f_j$ (orange), but not of the true bounding box $\hat{\cB}^f_j$ (blue).
    }
    \label{fig:example_background}
\end{figure}

\vspace{-1em}
\paragraph{Background Loss}
The background loss term $L_{BG}$ penalises any probability mass that the detector incorrectly assigned to pixels outside the ground-truth bounding box. It is formally defined as
\begin{equation}
    L_{BG}(\cG^f_i, \cD^f_j) = - \frac{1}{|\hat{\cS}^f_i|} \sum_{\vx \in \cV^f_{ij}} \log((1 - P(\vx \in \cS^f_j))),
    \label{eq:bgloss}
\end{equation}
which is the sum of negative log-probabilities of all pixels in the set $\cV^f_{ij} = \{ \cS^f_j - \hat{\cB}^f_i \}$, i.e. pixels that are part of the detection, but not of the true bounding box. A visualisation of this evaluation region $\cV^f_{i,j}$ is shown in Figure~\ref{fig:example_background}. 
Note that we average over $|\hat{\cS}^f_i|$ rather than $|\cV^f_{i,j}|$ to ensure that foreground and background losses are scaled equivalently, measuring the loss incurred per ground-truth pixel the detection aims to describe.
The background loss term is minimised if all pixels outside the ground-truth bounding box are assigned a spatial probability of zero.

\subsection{Label Quality}
While spatial quality measures how well the detection describes \textit{where} the object is within the image, label quality $Q_L$ measures how effectively a detection identifies \textit{what} the object is.
We define $Q_L$ as the probability estimated by the detector for the object's ground-truth class.
Note that this is irrespective of whether this class is the highest ranked in the detector's probability distribution.
Unlike with mAP, this value is explicitly used to influence detection quality rather than just for ranking detections regardless of actual label probability.
We define label quality as:
\begin{equation}
    Q_L(\cG^f_i, \cD^f_j) = \textbf{l}^f_j(\hat{c}^f_i).
\end{equation}

\subsection{Pairwise PDQ (pPDQ)}\label{sec:def:pdq:ppdq}

The pairwise PDQ (pPDQ) between a detection $\cD^f_j$ and a ground-truth object $\cG^f_i$ in frame $f$ is the geometric mean of the spatial quality and label quality measures $Q_{S}$ and $Q_{L}$:
\begin{equation}
    \operatorname{pPDQ}(\cG^f_i, \cD^f_j) = \sqrt{Q_{S}(\cG^f_i, \cD^f_j) \cdot Q_{L}(\cG^f_i, \cD^f_j)}.
    \label{eq:ppdq}
\end{equation}
Using the geometric mean requires both components to have high values for a high pPDQ score, and is zero if either component reaches zero.
Notice that it is also possible to use a weighted geometric mean for applications where the spatial or label quality component is more important.

\subsection{Assignment of Optimal Detection-Object Pairs}\label{sec:def:pdq:assign}
It is important that, for every frame, each detection is matched to, at most, one ground-truth object and vice versa.
This is also done for mAP, but it utilises a greedy assignment process based upon label confidence ranking, rather than ensuring that the optimal assignment takes into account both the spatial and label aspects of the detection.
To mitigate this problem, we use our proposed pPDQ score in \eqref{eq:ppdq} between possible detection-object pairings to determine the optimal assignment through the Hungarian algorithm~\cite{kuhn1955hungarian}.
This provides the optimal assignment between two sets of information which produce the best total pPDQ score. 

Using assignments from the Hungarian algorithm, we store the pPDQs for all non-zero assignments in the $f$-th frame in a vector $\textbf{q}^f = [q^f_1, q^f_2, q^f_3, ..., q^f_{N^f_{TP}}]$ where $N^f_{TP}$ is the number of non-zero (true positive) assignments within the $f$-th frame.
Note that these ``true positive'' detections are not ones which are considered 100\% accurate as is done for AP-based measures.
Instead these are detections which, even marginally, describe the ground-truth object they are matched with and provide a non-zero pPDQ.
If the pPDQ from an optimal assignment is zero, there is no association between the ground-truth object and detection.
This occurs when either a ground-truth object is undetected (false negative) or a detection does not describe an object (false positive). 
We also record the number of false negatives and false positives for each frame, expressed formally as $N^f_{FN}$ and $N^f_{FP}$ respectively, to be used in our final evaluation.
After obtaining $\textbf{q}^f$, $N^f_{TP}$, $N^f_{FN}$, and $N^f_{FP}$ for each frame, the PDQ score can be calculated.

\subsection{PDQ Score}\label{sec:def:pdq:final_pdq}

The final PDQ score across a set of ground-truth objects $\cG$ and detections $\cD$ is the total pPDQ for each frame divided by the total number of TPs, FNs and FPs assignments across all frames.
This can be seen as the average pPDQ across all TPs, FNs and FPs observed, which is calculated as follows:
\begin{equation}
    PDQ(\cG, \cD) = \frac{1}{\sum_{f =1}^{N_F} N^f_{TP} + N^f_{FN} + N^f_{FP}} \sum_{f =1}^{N_F}\sum_{i=1}^{N^f_{TP}} \textbf{q}^f(i),
\end{equation}
where $\textbf{q}^f(i)$ is the pPDQ score for the $i$-th assigned detection-object pair in the $f$-th frame.
This final PDQ score provides a consistent, probability-based measure, evaluating both label and spatial probabilities, that can determine how well a set of detections has described a set of ground-truth objects without the need for thresholds to determine complete success or failure of any given detection.

	\section{Evaluation of PDQ Traits}\label{sec:eval_pdq}
The previous section introduced PDQ, a new measure to evaluate the performance of detectors for \emph{probabilistic} object detection. PDQ has been designed with one main goal in mind: it should reward detectors that can accurately quantify both their spatial and label uncertainty. In this section, we are going to demonstrate that this goal has been met, by showing PDQ's behaviour in controlled experiments.
We show the most critical experiments here and more are provided in supplementary material.

\vspace{-1em}
\paragraph{PDQ Rewards Accurate Spatial Uncertainty} \label{sec:traits:spatial}
We perform a controlled experiment on the COCO 2017 validation dataset with a simulated object detector. For every ground truth object with true bounding box corners $\hat\vx_0$ and $\hat\vx_1$, the detector generates a detection with bounding box corners sampled as $\vx_0 \sim \cN(\hat\vx_0,\hat\vSigma)$ and $\vx_1 \sim \cN(\hat\vx_1,\hat\vSigma)$, with $\hat\vSigma = \operatorname{diag}(\hat{s}^2, \hat{s}^2)$. We vary the value of $\hat s^2$ throughout the experiments and refer to $\hat s^2$ as the detector's \textit{true} variance. Independent of the value of $\hat s^2$, the simulated detector expresses spatial uncertainty for each probabilistic detection with a different variance $\sigma^2$, which we refer to as the \emph{reported} variance. Each detection is assigned probability 1.0 for the \emph{true} label, corresponding to perfect classification.

When varying the values of $\hat{s}^2$ and $\sigma^2$ and evaluating the resulting detections, PDQ should reward when $\hat{s}^2$ is similar to $\sigma^2$, i.e. when the \emph{reported} spatial uncertainty is close to the \textit{true} spatial uncertainty that was used to sample the detection corners. When both \emph{reported} and \textit{true} spatial uncertainty are equal, PDQ should reach its peak performance. This would indicate that PDQ does indeed reward the accurate estimation of spatial uncertainty. 

\begin{figure}
    \centering
    \includegraphics[width=0.9\linewidth]{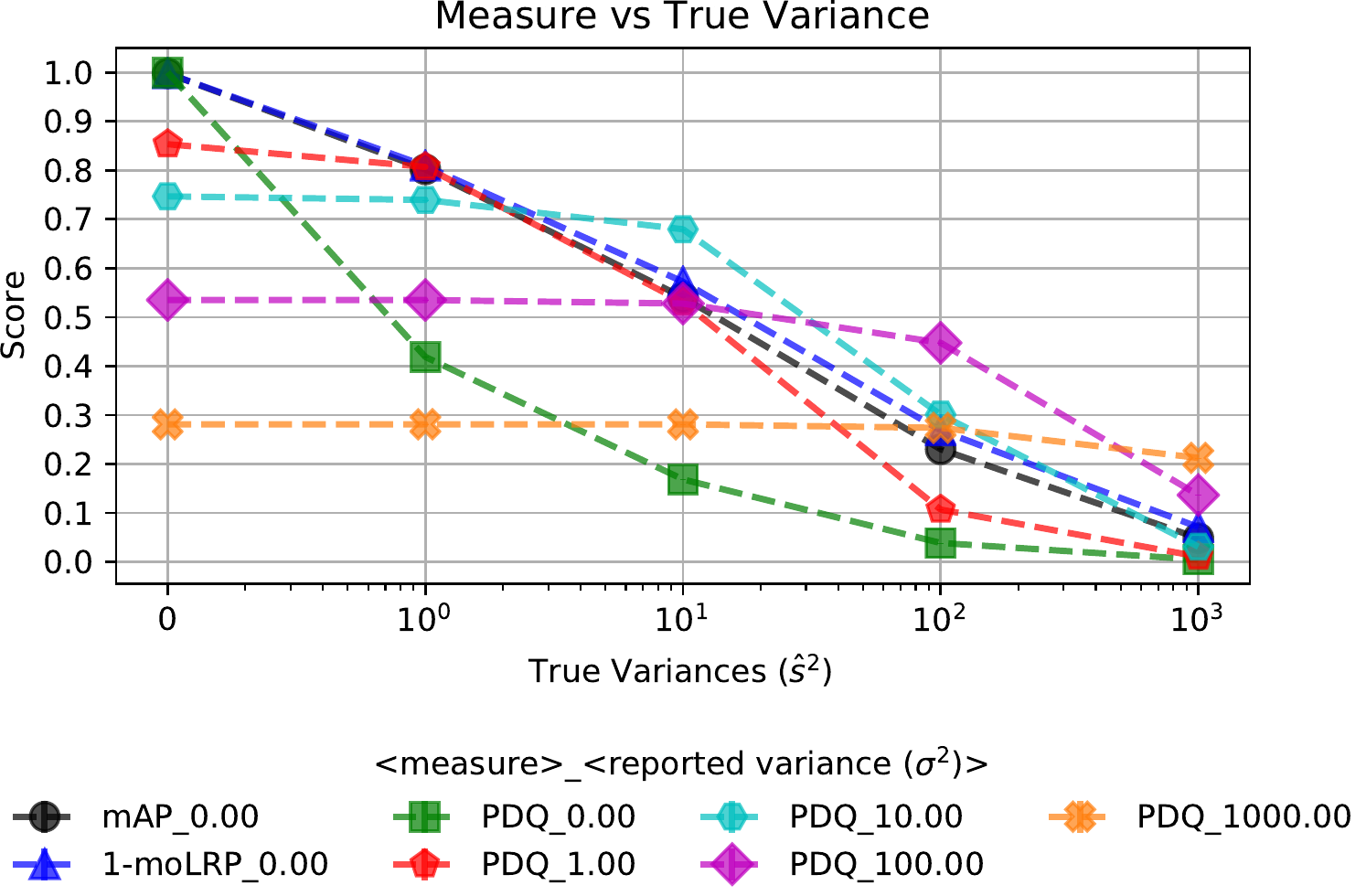}
    \includegraphics[width=0.9\linewidth]{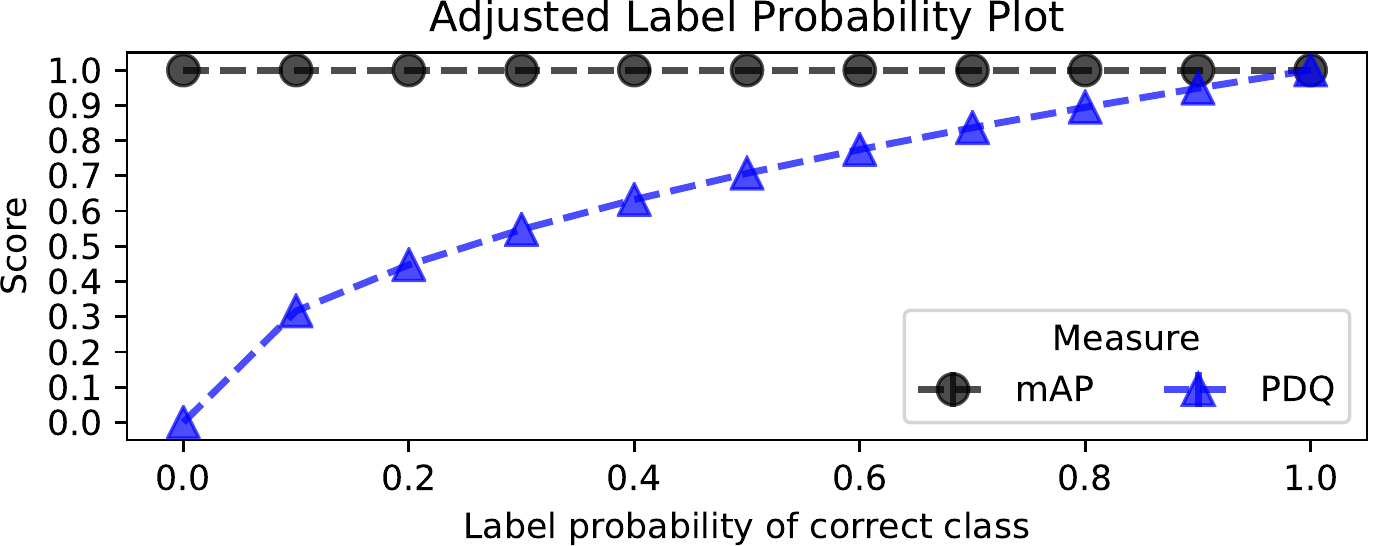}
      \caption{Top: PDQ rewards detectors that accurately evaluate their true spatial uncertainty. Bottom: PDQ explicitly evaluates label uncertainty, in contrast to mAP. See Section \ref{sec:traits:spatial} for explanation of the experiments.}
    \label{fig:eval_traits}
\end{figure}

Figure~\ref{fig:eval_traits} confirms this conjecture. We repeated the experiment described above 20 times, evaluating on all objects in the 5,000 images of the COCO 2017 validation set. Every line corresponds to a detector with a different \emph{reported} variance $\sigma^2$. The \textit{true} variance $\hat s^2$ varies along the x axis. We can see that the for each value of $\hat s^2$, simulated detectors with $\sigma^2 = \hat s^2$ give the best performance.

\vspace{-1em}
\paragraph{PDQ Explicitly Evaluates Label Uncertainty}
We perform a controlled experiment in a simulated scenario where a single object is detected by a single detection with perfect spatial quality.
We vary the detection's reported label probability for the correct class and ensure that it always remains the dominant class in the probability distribution. The resulting PDQ and mAP scores are compared in Figure~\ref{fig:eval_traits}. We observe that PDQ is affected by the label probability of the correct class via its label quality term. This is in contrast to mAP which uses label probability only to determine the dominant class and for ranking detection matches.

\section{Evaluation of Object Detectors}\label{sec:eval_real_detectors}
In this section we evaluate a number of state-of-the-art conventional detectors and the recently proposed probabilistic object detector MC-Dropout SSD~\cite{miller2018dropout} that is based on Monte Carlo Dropout. 
We compare the ranking of all tested detectors using PDQ and its components, as well as the established measures mAP and moLRP~\cite{oksuz2018localization}, and discuss our most important observations and gained insights.

\begin{table*}[tb]
\caption{PDQ-based Evaluation of Probabilistic and Non-Probabilistic Object Detectors. Legend: mLRP = $1 - \text{moLRP}$, Sp = Spatial Quality, Lbl = Label Quality, FG = Foreground Quality ($\exp(-L_{FG})$), BG = Background Quality ($\exp(-L_{BG}$), TP = True Positives, FP = False Positives, FN = False Negatives. pPDQ, Sp, Lbl, FG and BG averaged over all TP.}
\rowcolors{3}{}{lightgray}
\begin{tabular}{@{}lrrrrrrrrrrr@{}}
\toprule
Approach ($\tau$) & mAP & mLRP & PDQ & pPDQ & Sp & Lbl & FG & BG & TP & FP & FN \\ 
 & (\%) & (\%) & (\%) & (\%) & (\%) & (\%) & (\%) & (\%) & & & \\
\midrule
probFRCNN (0.5) & 35.5 &  32.2 & \textbf{28.4} & \textbf{56.7} & \textbf{45.0} & 90.7 & \textbf{77.8} & \textbf{60.7} & 23,434 & 10,016 & 13,347 \\
MC-Dropout SSD (0.5) \cite{miller2018evaluating} & 15.8 & 15.6 & 12.8 & 47.3 & 39.9 & 74.0 & 73.1 & 57.3 & 10,510 & \textbf{2,165} & 26,271 \\	
MC-Dropout SSD (0.05) \cite{miller2018evaluating} & 19.5 & 16.6 & 1.3 & 26.1 & 27.3 & 35.9 & 60.1 & 46.2 & 24,843 & 461,074 & 11,938 \\	
SSD-300 (0.5) \cite{liu2016ssd} & 15.0 & 14.3 & 3.9 & 18.1 & 9.7 & 80.2 & 57.5 & 25.1 & 8,999 & 4,746 & 27,782	 \\	
SSD-300 (0.05) \cite{liu2016ssd} & 19.3 & 16.0 & 0.6 & 9.7 & 6.4 & 40.2 & 38.1 & 32.3 & 21,961 & 324,067 & 14,820	 \\ 
YOLOv3 (0.5) \cite{redmon_yolov3:_2018} & 29.7 & 30.8 & 5.7 & 14.6 & 6.2 & \textbf{95.8} & 52.2 & 20.4 & 17,390 & 7,728 & 19,391	 \\	
YOLOv3 (0.05) \cite{redmon_yolov3:_2018} & 30.1 & 27.7 & 3.3 & 12.2 & 5.1 & 92.8 & 44.6 & 22.9 & 23,447 & 50,074 & 13,334	 \\	
FRCNN R (0.5) \cite{jjfaster2rcnn} & 32.8 & 29.1 & 6.7 & 19.1 & 10.3 & 88.8 & 62.2 & 23.6 & 19,930 & 20,044 & 16,851	 \\
FRCNN R (0.05) \cite{jjfaster2rcnn} & 34.3 & 29.1 & 3.0 & 17.1 & 9.5 & 78.5 & 57.8 & 25.1 & 23,081 & 93,141 & 13,700	 \\
FRCNN R+FPN (0.5) \cite{massa2018mrcnn} & 34.6 & 31.2 & 11.8 & 27.1 & 16.9 & 86.5 & 60.6 & 35.7 & 22,537 & 14,706 & 14,244 \\
FRCNN R+FPN (0.05) \cite{massa2018mrcnn} & 37.0 & 30.4 & 4.2 & 23.1 & 15.8 & 69.5 & 54.4 & 38.7 & 29,326 & 123,511 & 7,455 \\
FRCNN X+FPN (0.5) \cite{massa2018mrcnn} & 37.4 & \textbf{32.7} & 11.9 & 27.9 & 17.6 & 88.2 & 60.8 & 36.8 & 24,523 & 20,444 & 12,258 \\
FRCNN X+FPN (0.05) \cite{massa2018mrcnn} & \textbf{39.0} & 32.1 & 4.4 & 24.8 & 16.7 & 74.4 & 55.6 & 39.1 & \textbf{29,922} & 130,009 & \textbf{6,859} \\
\bottomrule
\end{tabular}
\label{tab:evaluation}
\end{table*}

\subsection{Experimental Set-up}
\paragraph{Evaluated Detectors}
The state-of-the-art conventional object detectors evaluated were SSD-300~\cite{liu2016ssd}, YOLOv3~\cite{redmon_yolov3:_2018}, FasterRCNN with ResNet backbone (FRCNN R)~\cite{jjfaster2rcnn}, FasterRCNN with ResNet backbone and feature pyramid network (FPN) (FRCNN R+FPN)~\cite{massa2018mrcnn}, and FasterRCNN with ResNeXt backbone and FPN (FRCNN~X+FPN)~\cite{massa2018mrcnn}.
To evaluate these conventional detectors with PDQ, we set $P(\vx \in \cS^f_j) = 1 - \epsilon$ for all pixels $\vx$ inside the provided standard bounding box, and $\epsilon$ for all pixels outside, when performing the calculations in equations (\ref{eq:fgloss}) and (\ref{eq:bgloss}), with $\epsilon=10^{-14}$ to avoid infinite loss.

In addition to conventional object detectors, we evaluate a probabilistic MC-Dropout object detector based on the work by Miller et al.~\cite{miller2018dropout, miller2018evaluating}. We follow the established implementation~\cite{miller2018evaluating}, where Monte Carlo Dropout~\cite{gal2015bayesian} is utilised in a SSD-300 object detector~\cite{liu2016ssd} with two dropout layers inserted and activated during both training and testing. Each image is tested with 20 forward passes through the network with randomised dropout masks to obtain samples of detections.
The recommended merging strategy was used to cluster these samples~\cite{miller2018evaluating}, namely a BSAS clustering method~\cite{theodoridis03bsas} with spatial affinity IoU and label affinity `same label' (we found an IoU threshold of 0.7 performed better than the 0.95 threshold recommended in~\cite{miller2018evaluating}). Final probabilistic detections were obtained by averaging sample label probability distributions and estimating $\cN_0$ and $\cN_1$ from the average and covariance of sample bounding boxes.

We furthermore modify a FasterRCNN with ResNeXt backbone and feature pyramid network to approximate probabilistic detections. We achieve this by the following process: for every detection surviving the normal non-maximum suppression, we find all of the suppressed detections that have an IoU of above 0.75 with the surviving detections and cluster them (including the survivors). We then calculate the Gaussian corner mean and covariances of each cluster, weighted by the detection's winning label confidences. We denote this method as probFRCNN in Table~\ref{tab:evaluation}.

\vspace{-1em}
\paragraph{Evaluation Protocol and Datasets}
Evaluation was performed on the 5,000 images of the MS COCO 2017 validation set~\cite{lin_microsoft_2014}, after all detectors have been trained or finetuned on the 2017 training dataset. During the evaluation, we ignored all detections with the winning class label probability below a threshold $\tau$. We compare the effect of this  process for $\tau = 0.5$ and $0.05$.

\subsection{Insights}
Table \ref{tab:evaluation} presents the results of our evaluation, comparing PDQ and its components with mAP and moLRP. 
From these results we observe the following:

\vspace{-1em}
\paragraph{1. PDQ exposes the performance differences between probabilistic and non-probabilistic object detectors.}
When evaluating using mAP or moLRP, both SSD-300~\cite{liu2016ssd} and FasterRCNN with ResNeXt and FPN~\cite{massa2018mrcnn}, and their respective probabilistic variants (MC-DropoutSSD~\cite{miller2018evaluating} and probFRCNN) show very similar performance. However, evaluating with PDQ reveals their performance differences in terms of probabilistic object detection: both probabilistic variants perform significantly better than their non-probabilistic counterparts. This is especially true for their overall spatial quality and its foreground and background quality components. Comparing probFRCNN with MC-DropoutSSD, we found that probFRCNN achieved a higher PDQ score, benefiting from its more accurate base network.

\vspace{-1em}
\paragraph{2. PDQ reveals differences in spatial and label quality.}
Since PDQ comprises meaningful components, it allows a detailed analysis of how well detectors perform in terms of spatial and label quality. For example, in Table~\ref{tab:evaluation}
we observe that the YOLOv3 detector achieves the highest label quality (95.8\%/92.8\% for $\tau=0.5/0.05$), but the worst spatial quality (6.2\%/5.1\%) of all tested detectors. This gives important insights into worthwhile directions of future research, suggesting YOLO can be more trusted to understand \textit{what} an object is than other detectors but is less reliable in determining precisely \textit{where} that object is.

\vspace{-1em}
\paragraph{3. Probabilistic localisation performance of existing object detectors is weak.} 
Spatial quality in PDQ measures how well detectors probabilistically localise objects in an image.
Conventional object detectors assume full confidence in their bounding box location and achieve low spatial qualities between $5.1\%$ and $17.6\%$, indicating they are spatially overconfident.
Since conventional object detectors have comparatively high label qualities, we conclude that for probabilistic object detection tasks where spatial uncertainty estimation is important, improving the localisation performance and the estimation of spatial uncertainty has the biggest potential of improving performance.

\begin{figure}[t]
    \centering
    \begin{subfigure}[b]{0.42\linewidth}
        \includegraphics[width=\textwidth]{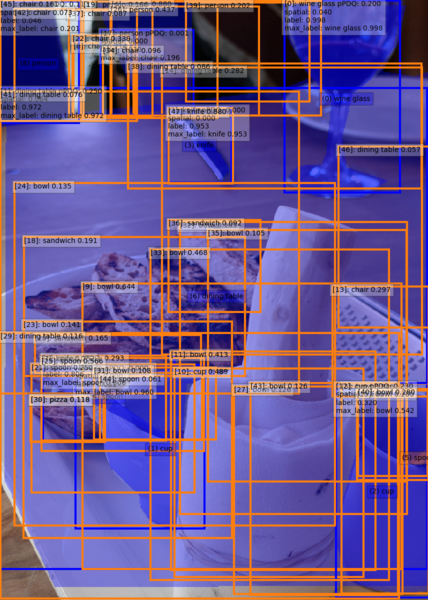}
        \caption{FRCNN X+FPN (0.05)}
        \label{fig:resnext_0.05}
    \end{subfigure}
    \begin{subfigure}[b]{0.42\linewidth}
        \includegraphics[width=\textwidth]{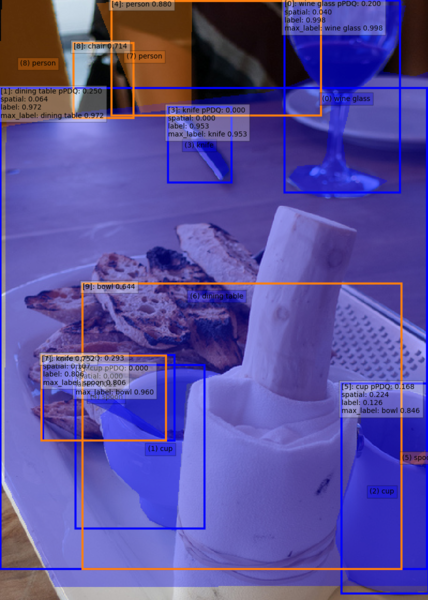}
        \caption{FRCNN X+FPN (0.5)}
        \label{fig:resnext_0.5}
    \end{subfigure}
    \caption{Visualisation of all TPs (blue segmentation mask and corresponding BBox), FPs (orange BBox), and FNs (orange segmentation mask) as defined by PDQ for FRCNN X+FPN with $\tau=0.5$ (a) and $\tau=0.05$ (b).
    We see here that a lower $\tau$ leads to far more FPs that are strongly penalised by PDQ but are largely ignored under mAP.}
    \label{fig:fp_comparison}
\end{figure}

\vspace{-1em}
\paragraph{4. PDQ does not obscure false positive errors.}
Unlike mAP and moLRP, PDQ explicitly penalises a detector for spurious (false positive) detections, as well as for missed (false negative) detections. We observe that decreasing the label threshold $\tau$ and consequently massively increasing the number of false positive detections (see Fig.~\ref{fig:fp_comparison} for an example) actually increases mAP, 
and does not tend to affect moLRP much. In contrast, PDQ scores decrease significantly. PDQ is designed to evaluate systems for application in real-world systems and does not filter detections based on label ranking or calculating the  optimal threshold $\tau$. It involves \emph{all} reported detections in its analysis.

	\section{Conclusions and Future Work}\label{sec:conclusion}
We introduced Probabilistic Object Detection, a challenging new task that is highly relevant for domains where accurately estimating the spatial and semantic uncertainties of the detections is of high importance such as embodied AI (such as robotics, autonomous systems, driverless cars), and medical imaging. To foster further research in this direction, we introduced the probability-based detection quality (PDQ) measure which explicitly evaluates both spatial and label uncertainty.

PDQ is not meant to \emph{replace} mAP, but to \emph{complement} it. Both evaluation measures are designed for two \emph{different} problems. While mAP has been the established performance measure for conventional object detection, we developed PDQ specifically for the new task of \emph{probabilistic} object detection.

After evaluating a range of object detectors, including the first emerging probabilistic object detector in Section~\ref{sec:eval_real_detectors}, we are confident that PDQ is a useful performance measure that can guide and inform the research of even better probabilistic object detectors in the future.
In future work we will investigate how to train object detectors to directly optimise for PDQ by incorporating it into the training loss function. 
The concept of probabilistic object detection can also be easily extended to Probabilistic \emph{Instance Segmentation} where each pixel would contain a probability of belonging to a certain object instance, along with a label distribution.

	{\small
		\bibliographystyle{ieee}
\bibliography{refs}
	}

\section*{Appendix Overview}
In this appendix we provide supplementary material and analysis that was not included in the main paper due to space restraints.
This appendix is organized as follows:

\begin{enumerate}[A.]
	\item PDQ Qualitative Examples.
	\item Evaluation of PDQ traits.
	\item Traditional Measures Obscuring False Positives.
	\item Definition of mAP.
\end{enumerate}

\section*{A. PDQ Qualitative Examples}
We provide qualitative results for detectors tested on COCO data in Section 7 of the main paper.
Specifically, in this section we visualise results from SSD-300~\cite{liu2016ssd}, YOLOv3~\cite{redmon_yolov3:_2018}, Faster RCNN with ResNext backbone and a feature pyramid network (FRCNN~X+FPN)~\cite{massa2018mrcnn}, and the probabilistic MC-Dropout SSD detector based on the work by Miller et al.~\cite{miller2018dropout, miller2018evaluating}.
Unless otherwise stated, results shown are for detectors using a label confidence threshold of 0.5.

Using the detection-object pairing assignment from PDQ as outlined in section 5.4 of the main paper, we are able to provide visualisations outlining the true positives (TPs), false positives (FPs) and false negatives (FNs) present in a given image, as was done in Figure 6 of the main paper.
In these visualisations we show TPs as blue segmentation masks and boxes, FPs as orange boxes, and FNs as orange segmentation masks.
We also provide a way to visualise spatially probabilistic detections using \textbf{ellipses} in the top-left and bottom-right corners, showing the contours of the Gaussian corners at distances of 1, 2 and 3 standard deviations.
For conventional detectors, there are no ellipses as they provide no spatial uncertainty.
Because we know the optimal assignment, as mentioned in the main paper, we can extract pairwise quality scores between TPs.
In our visualisations we provide pPDQ, spatial quality and label quality scores for all TP detections in a text box at the top-left corner of the detection box.

Using visualisations of this form enables us to qualitatively reinforce some of the findings from the main paper in the following three subsections.
Firstly, we see again how the number of false positives under PDQ increases with lower label confidence thresholds (despite such detections getting higher mAP scores).
Secondly, we get to observe the effect of spatial uncertainty estimation and how this effects spatial quality scores for different detections.
Thirdly, we can visually show the high label quality but poorer localisation achieved by YOLOv3 when compared to FRCNN X+FPN.

\subsection*{A.1. Increased False Positives with Lower Label Confidence Threshold}
\begin{figure*}[t]
    \centering
    \begin{subfigure}[t]{\linewidth}
    \centering
    \begin{subfigure}[b]{0.2\linewidth}
    \includegraphics[width=\textwidth]{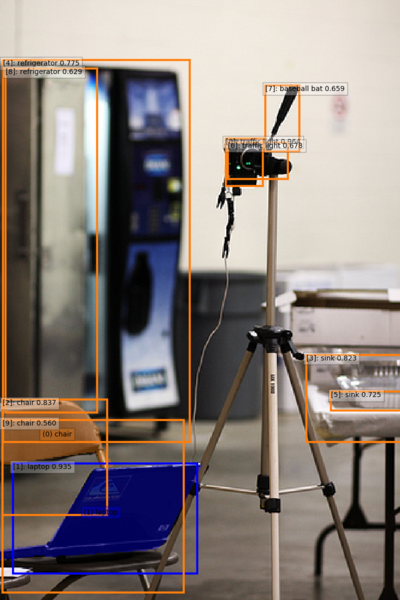}
    \end{subfigure}
    \begin{subfigure}[b]{0.2\linewidth}
    \includegraphics[width=\textwidth]{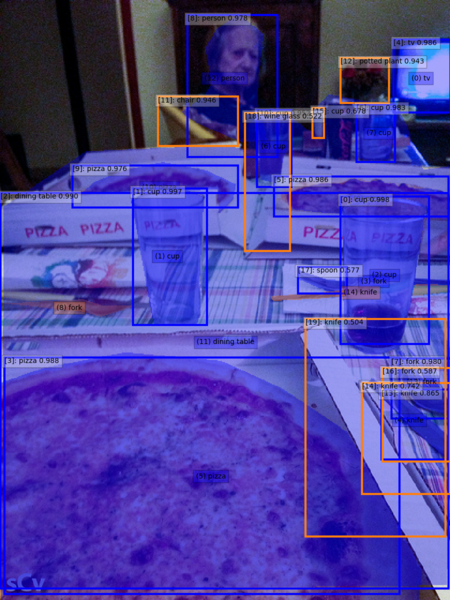}
    \end{subfigure}
    \begin{subfigure}[b]{0.2\linewidth}
    \includegraphics[width=\textwidth]{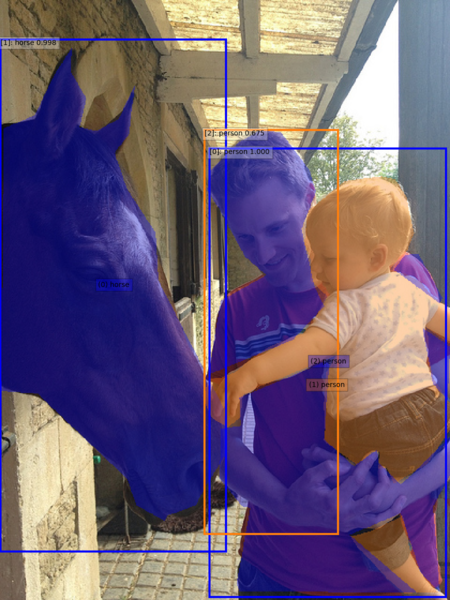}
    \end{subfigure}
    \begin{subfigure}[b]{0.2\linewidth}
    \includegraphics[width=\textwidth]{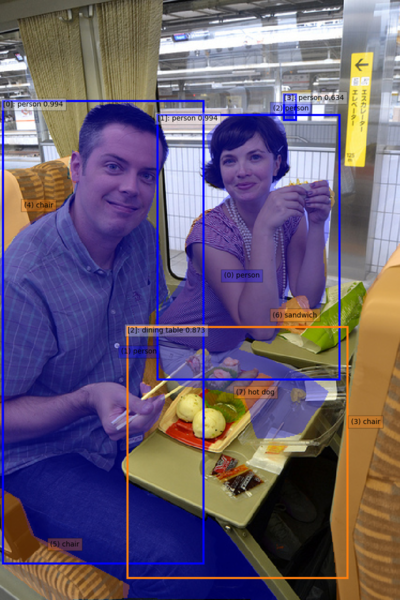}
    \end{subfigure}
    \caption{0.5 Label Threshold}
    \end{subfigure}
    \begin{subfigure}[t]{\linewidth}
    \centering
    \begin{subfigure}[b]{0.2\linewidth}
    \includegraphics[width=\textwidth]{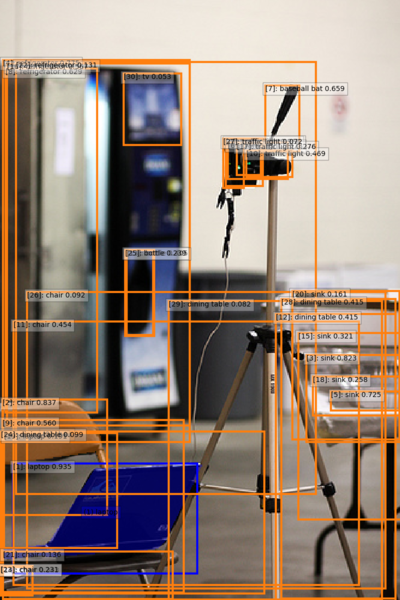}
    \end{subfigure}
    \begin{subfigure}[b]{0.2\linewidth}
    \includegraphics[width=\textwidth]{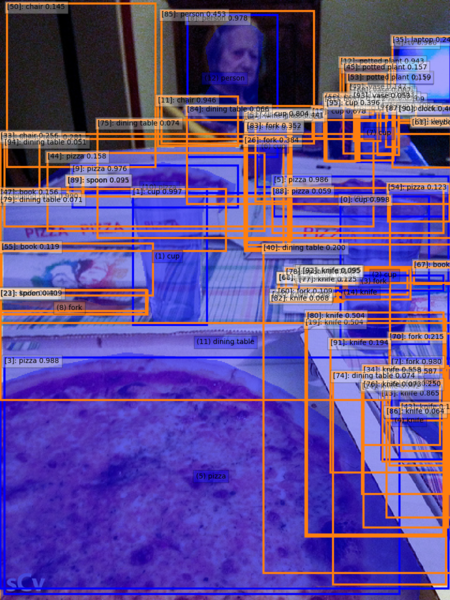}
    \end{subfigure}
    \begin{subfigure}[b]{0.2\linewidth}
    \includegraphics[width=\textwidth]{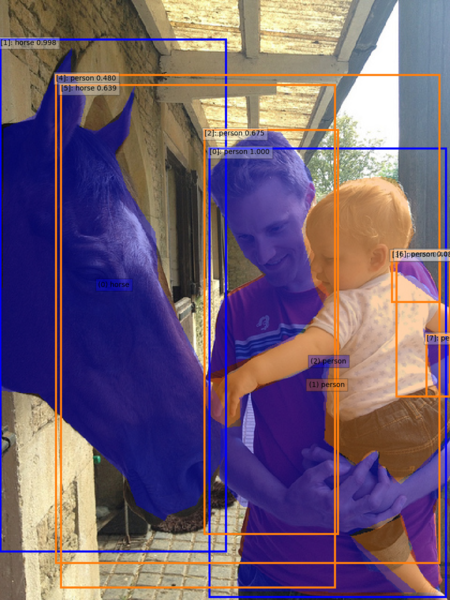}
    \end{subfigure}
    \begin{subfigure}[b]{0.2\linewidth}
    \includegraphics[width=\textwidth]{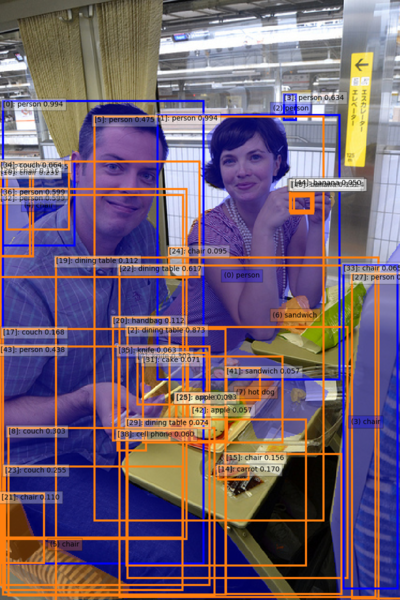}
    \end{subfigure}
    \caption{0.05 Label Threshold}
    \end{subfigure}
    \caption{Detections from FRCNN X+FPN at label confidence thresholds of 0.5 (a) and 0.05 (b) as evaluated by PDQ. We see more false positives (orange boxes) under PDQ with 0.05 despite 0.05 giving higher mAP scores as shown in the main paper.}
    \label{fig:qual_thresholds}
\end{figure*}

Reinforcing the finding of the main paper, we show more examples for FRCNN X+FPN with label confidence thresholds of 0.5 and 0.05 respectively in Figure~\ref{fig:qual_thresholds}.
Note that because these images are rather cluttered, we omit the detailed quality information beyond the detection's maximum class label.
We see that the number of FPs (orange boxes) increases dramatically when the label confidence threshold is lowered to 0.05.

\subsection*{A.2. Spatial Uncertainty Estimation}
\begin{figure}
    \centering
    \begin{subfigure}[b]{0.9\linewidth}
    \includegraphics[width=\textwidth]{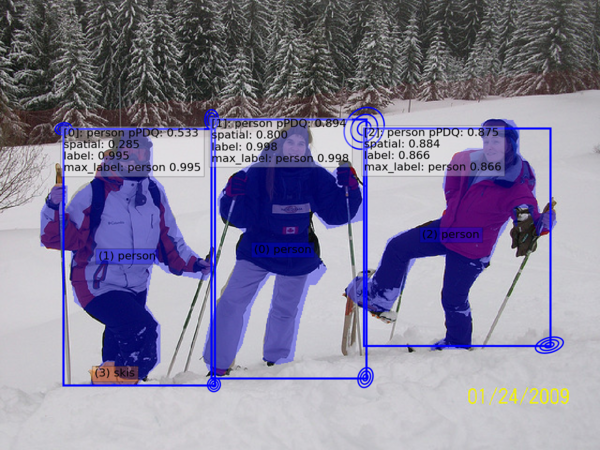}
    \caption{general}
    \label{fig:qual_spatial_uncertainty:general}
    \end{subfigure}
    \begin{subfigure}[b]{0.9\linewidth}
    \includegraphics[width=\textwidth]{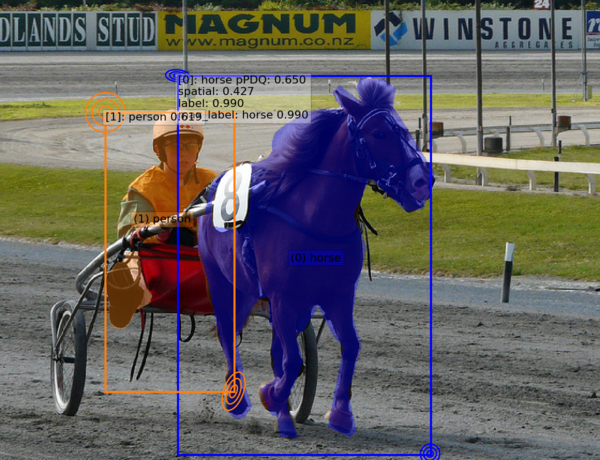}
    \caption{over-confident}
    \label{fig:qual_spatial_uncertainty:over-confident}
    \end{subfigure}
    \begin{subfigure}[b]{0.9\linewidth}
    \includegraphics[width=\textwidth]{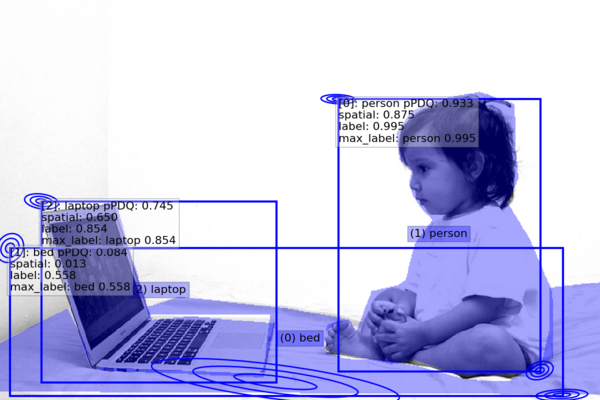}
    \caption{under-confident}
    \label{fig:qual_spatial_uncertainty:under-confident}
    \end{subfigure}
    \caption{Visualisation of MC-Dropout SSD detections as analysed by PDQ. Ellipses represent spatial uncertainty. In (a) we see a general case where individuals have better or worse spatial quality dependant on uncertainty estimation. In (b) we see a detection with uncertainty which is still over-confident and misses the person. In (c) we see an under-confident detection around the laptop.}
    \label{fig:qual_spatial_uncertainty}
\end{figure}

We show some examples from the MC-Dropout SSD detector to highlight the effect that spatial uncertainty has on both spatial quality and overall pPDQ in Figures~\ref{fig:qual_spatial_uncertainty} and~\ref{fig:qual_dropout_compare}.

Figure~\ref{fig:qual_spatial_uncertainty} shows the effect that spatial uncertainty estimation has on the spatial quality of PDQ.
In Figure~\ref{fig:qual_spatial_uncertainty:general} we see the spatial quality vary between three people based upon uncertainty estimation.
The left-most person has the poorest spatial quality as the box misses part of his entire arm, goes too far below their feet, and yet has very little spatial uncertainty in it's detection, scoring a spatial quality of only 28.5\%.
This is in comparison to the right-most person who has a detection with some uncertainty to the top, left, and right of the box, matching where there is the most error in the detection itself.
This leads to a much higher spatial quality of 88.4\%.

In Figure~\ref{fig:qual_spatial_uncertainty:over-confident}, we see that simply adding spatial uncertainty is not enough to guarantee a good score and a TP detection.
We see the bottom of the detection box for the human is over-confident, leading to a FP detection.
Finally, in Figure~\ref{fig:qual_spatial_uncertainty:under-confident}, we see that the box around the laptop is nearly perfect and yet the right-most edge has high uncertainty.
By comparison, we see the person in the picture has a poorer base bounding box but appears to have a more reasonable estimate of it's uncertainties.
Comparing spatial quality scores, we see that despite it's better base bounding box, the spatial quality of the laptop is only 65\% compared the person's spatial quality of 87.5\%.
This drop in spatial quality is due to the high spatial uncertainty expressed by the laptop detection.

\subsection*{A.3. MC Dropout Vs SSD}
\begin{figure}
    \centering
    \begin{subfigure}[b]{0.83\linewidth}
    \includegraphics[width=\textwidth]{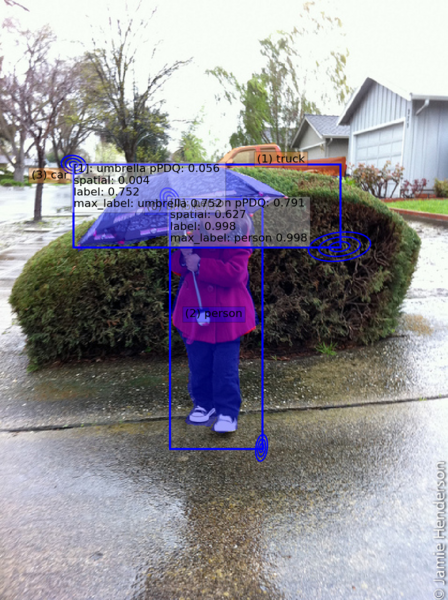}
    \caption{MC-Dropout SSD}
    \label{fig:qual_dropout_compare:dropout}
    \end{subfigure}
    \begin{subfigure}[b]{0.83\linewidth}
    \includegraphics[width=\textwidth]{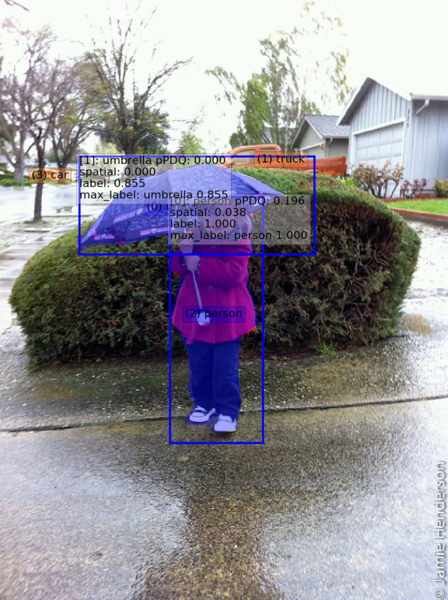}
    \caption{SSD-300}
    \label{fig:qual_dropout_compare:ssd}
    \end{subfigure}
    \caption{Comparison of MC-Dropout SSD to SSD-300. SSD-300 is shown to be spatially over-confident leading to low scores despite tighter boxes.}
    \label{fig:qual_dropout_compare}
\end{figure}

In the main paper, we showed that MC-Dropout SSD was able to achieve higher spatial quality,  and by extension pPDQ, than conventional detectors.
We show this visually in Figure~\ref{fig:qual_dropout_compare}, comparing detections from MC-Dropout SSD to those of SSD-300.
Neither has tight detections around the person or umbrella, but SSD-300 boxes visually appear tighter.
However, SSD-300 detections are over-confident, expressing no spatial uncertainty and attaining spatial quality up to only 3.8\% found on the person.
In comparison, we see MC-Dropout SSD detections expressing uncertainty that  coincides with the innacuraccies of the detection.
This provides a spatial quality of up to 62.7\% found on the person.
Better pPDQ scores are seen for both objects with MC-Dropout.

\begin{figure*}
    \centering
    \begin{subfigure}{\linewidth}
    \centering
    \begin{subfigure}{0.4\linewidth}
        \includegraphics[width=\textwidth]{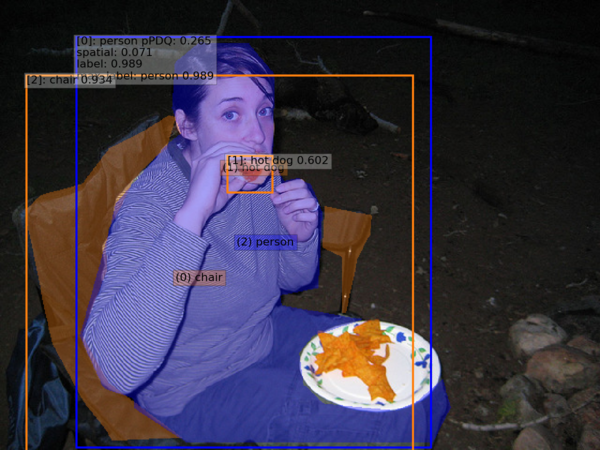}
    \end{subfigure}
    \begin{subfigure}{0.4\linewidth}
        \includegraphics[width=\textwidth]{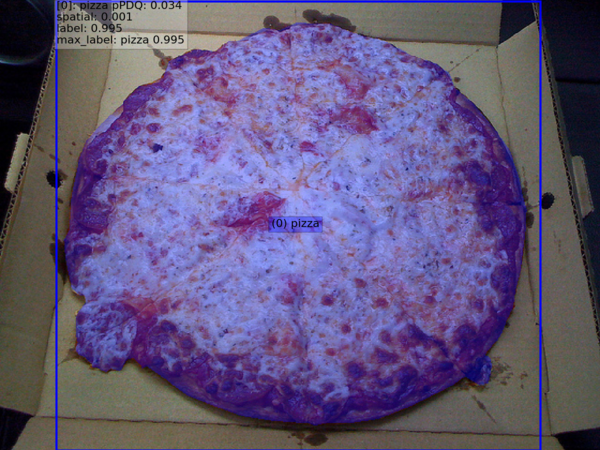}
    \end{subfigure}
    \caption{YOLOv3}
    \label{fig:qual_yolo:yolo}
    \end{subfigure}
    \begin{subfigure}{\linewidth}
    \centering
    \begin{subfigure}{0.4\linewidth}
        \includegraphics[width=\textwidth]{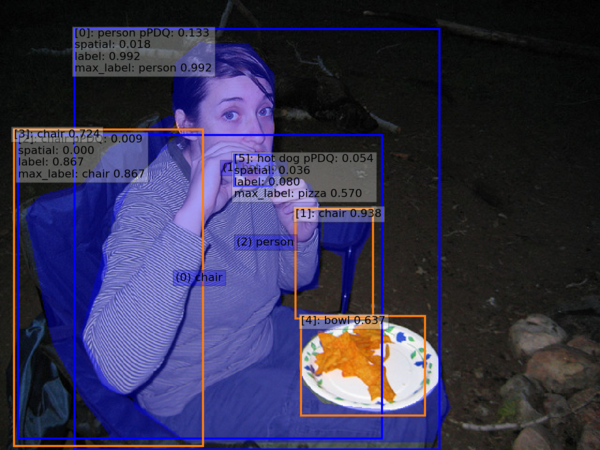}
    \end{subfigure}
    \begin{subfigure}{0.4\linewidth}
        \includegraphics[width=\textwidth]{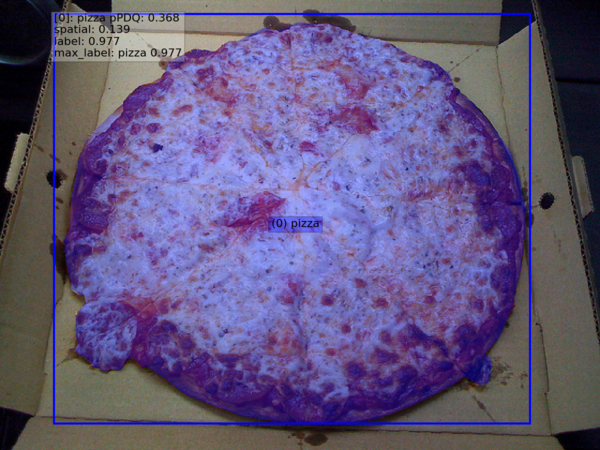}
    \end{subfigure}
    \caption{FRCNN X+FPN}
    \label{fig:qual_yolo:frcnn}
    \end{subfigure}
    \caption{Visualisation of YOLOv3 detections compared with FRCNN X+FPN.}
    \label{fig:qual_yolo}
\end{figure*}

\subsection*{A.4. YOLO Label Vs Spatial Quality}
In the experiments from the main paper, we showed that YOLOv3 achieves high label quality but comparatively low spatial quality when compared with other detectors such as FRCNN X+FPN.
In Figure~\ref{fig:qual_yolo} we visually compare YOLOv3 and FRCNN X+FPN results to qualitatively confirm this observation.

Examining Figure~\ref{fig:qual_yolo}, we see that in the left image YOLOv3 produces higher confidence detections for chair and hotdog than FRCNN X+FPN, but because their detections are over-confident and have poorer localisation, they are treated as FPs rather than TPs.
On the right, we see a more confident pizza detection from YOLOv3 but a poorer box localisation leading to spatial quality of 0.1\% compared to the 13.9\% spatial quality of FRCNN X+FPN (0.5).
This supports the observation from the main paper that YOLOv3 can have higher label quality than FRCNN detectors but tends to have a lower spatial quality due to poorer localisation.

\clearpage
\section*{B. Evaluation of PDQ Traits}
We demonstrate the characteristics of PDQ when compared with existing measures (mAP~\cite{lin_microsoft_2014} and moLRP~\cite{oksuz2018localization}) when responding to different types of imperfect detections, expanding upon what was covered in the main paper. 
Specifically, we examine the effect of spatial uncertainty, detection misalignment, label quality, missing ground-truth objects, and duplicate/false detections.
Throughout, we refer to standard detections with no spatial uncertainty as bounding box (BBox) detections and probabilistic detections with spatial uncertainty as probabilistc bounding box (PBox) detections.

\subsection*{B.1. Spatial Uncertainty}
We examine the effect of spatial uncertainty on BBox and PBox detections respectively.

\begin{figure}[t]
    \centering
    \includegraphics[width=0.8\linewidth]{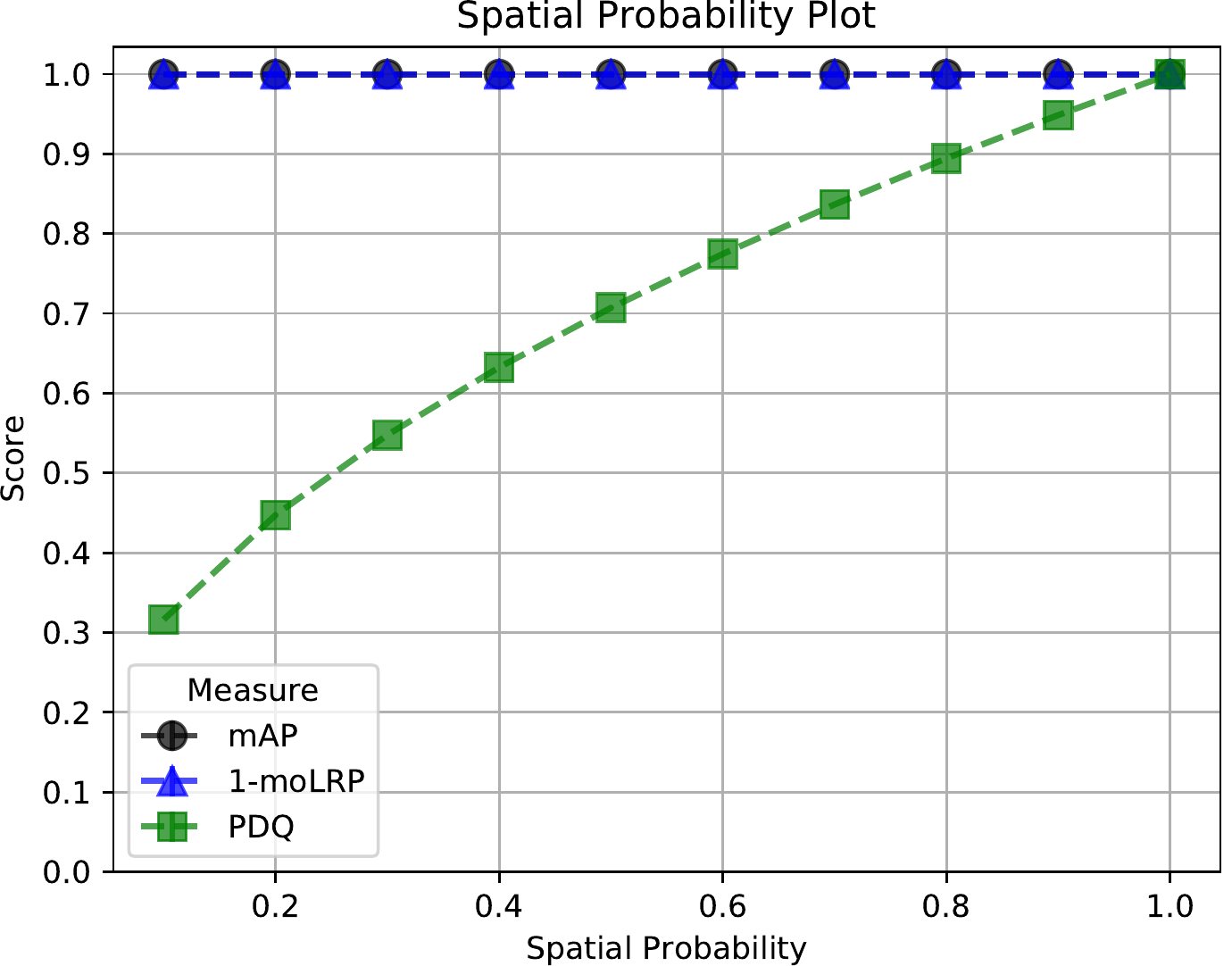}
    \caption{Evaluation of the effect of spatial probability on a perfectly aligned BBox. We see that unlike existing object detection measures, PDQ is effected by spatial probability changes.}
    \label{fig:eval_traits:spatial}
\end{figure}

\paragraph{BBox Spatial Uncertainty}
We evaluate a perfectly aligned BBox detection which has varying values of spatial probability for every pixel therein.
Whilst not a realistic type of detection, it allows for easy examination of the response from existing measures and PDQ to spatial probability variations.
The results are shown in Figure~\ref{fig:eval_traits:spatial}

This experiment shows that PDQ is gradually reduced by decreasing spatial certainty, whereas mAP and moLRP consistently consider the provided output to be perfect as they are not designed to measure uncertainty.

\begin{figure}[t]
    \centering
    \includegraphics[width=0.8\linewidth]{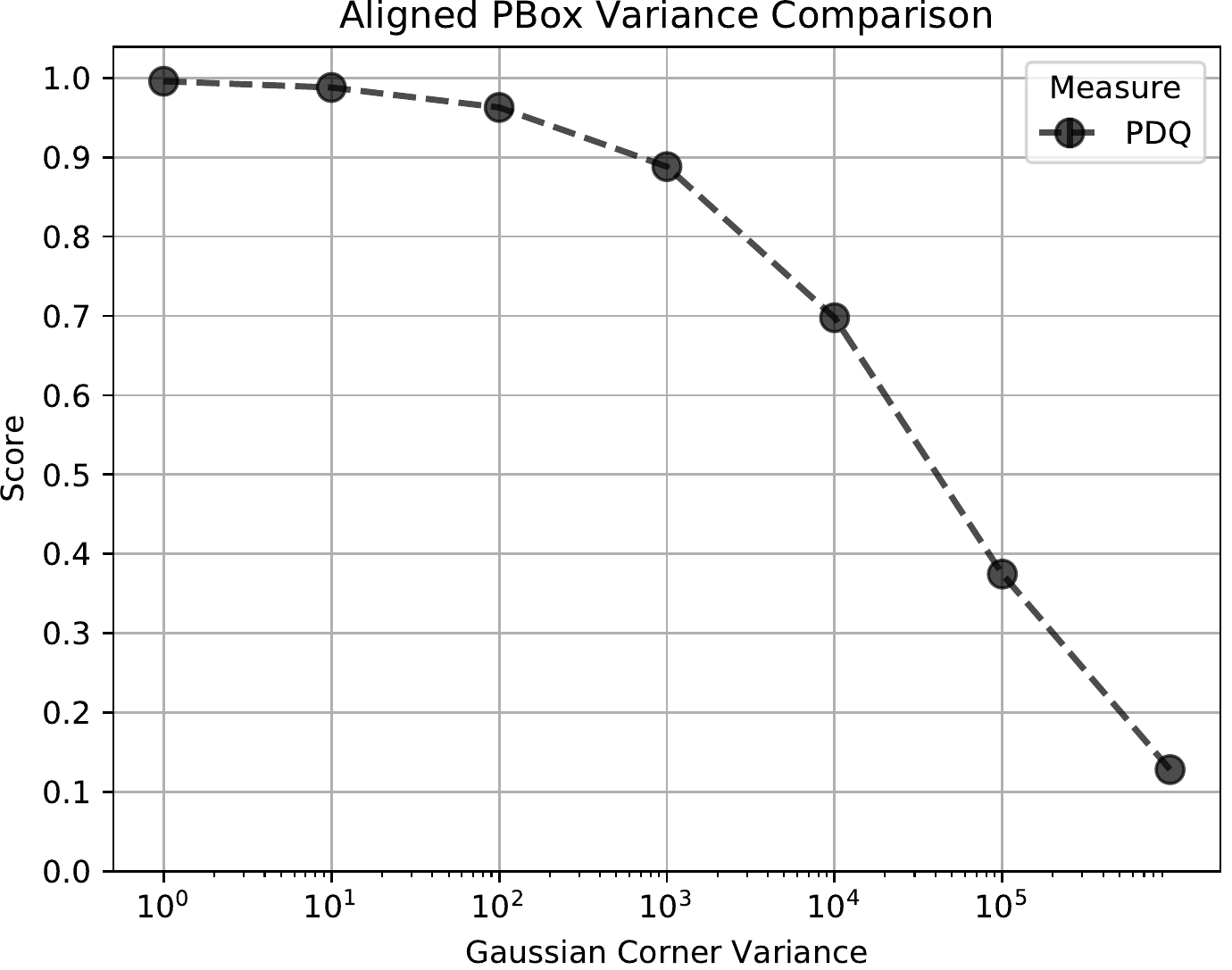}
    \caption{Plot showing the effect on PDQ of increasing variance, and by extension uncertainty, on perfectly aligned PBoxes. We see that for perfectly aligned detections, the score goes down the more uncertain the PBox detection is.}
    \label{fig:sup:cov}
\end{figure}

\paragraph{PBox Spatial Uncertainty}
To examine the effect of increasing spatial uncertainty on PDQ using PBoxes, we perform a test using a perfectly aligned PBox detection on a single object.
We consider a simple square-shaped 500 x 500 object centred in a 2000 x 2000 image.
PBox corner Gaussians are spherical and located at the corners of the object they are detecting. PBox reported variance for the corner Gaussians is varied to observe the effect of increased uncertainty.
The results of this test are shown in Figure~\ref{fig:sup:cov}.
We see a decline in PDQ with increased uncertainty demonstrating how PDQ penalises under-confidence.

\subsection*{B.2. Detection Misalignment}
\begin{figure}[t]
    \centering
    \includegraphics[width=0.7\linewidth]{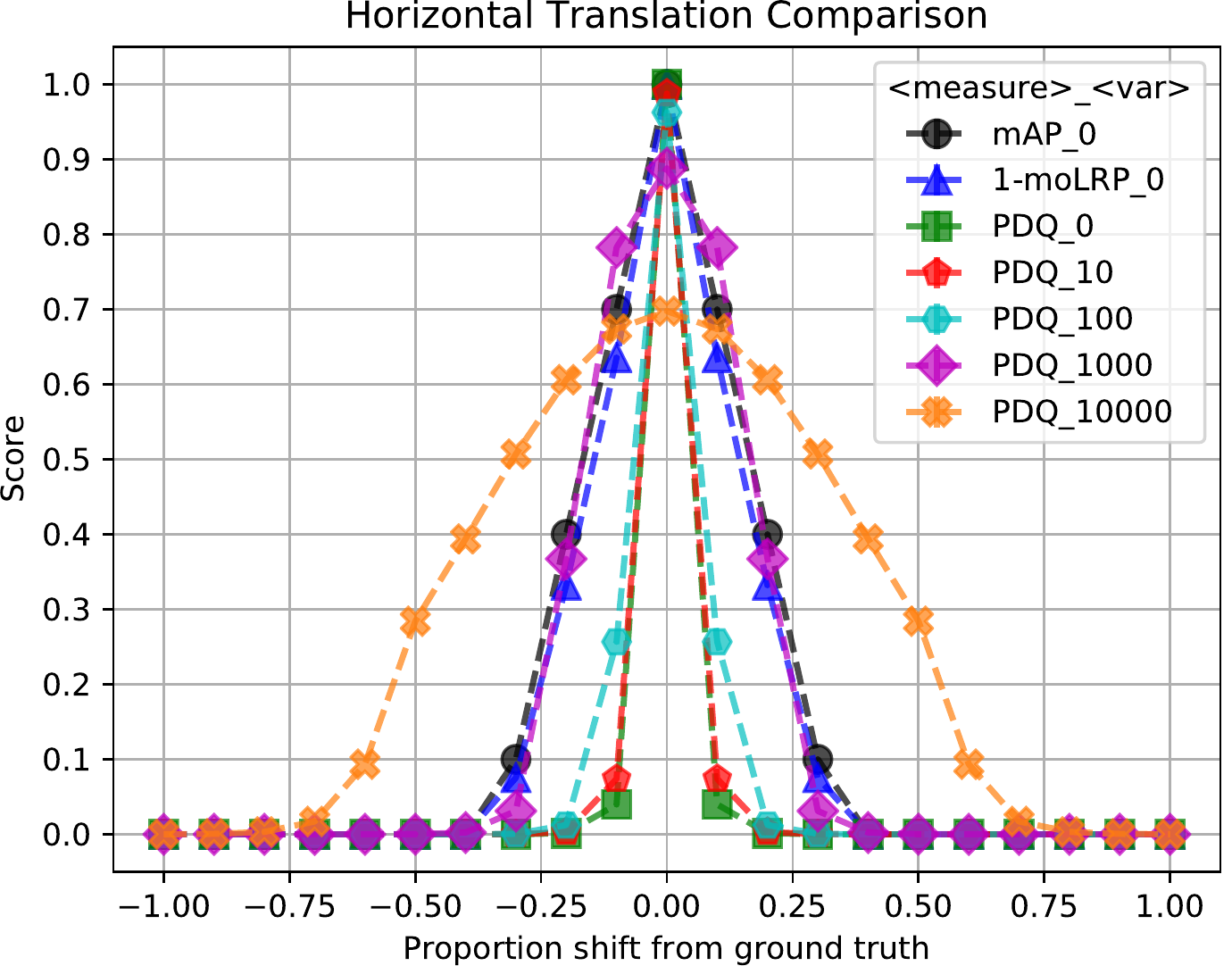}
    \caption{Evaluation of the effect of translation on mAP, moLRP, and PDQ scores. X-axis shows proportional shift of detection box either to the left (negative) or right (positive). Variance (\textit{var}) refers to the variance of corner Gaussians of the PBox detections. BBox is used when \textit{var} is zero. We see mAP and moLRP are lenient to BBox detections with no uncertainty when compared to PDQ and that PDQ is more lenient the more uncertain the detector is.}
    \label{fig:eval_traits:translation}
\end{figure}

We perform two experiments to analyse responses to misaligned detections.
These are translation error and scaling error.
\paragraph{Translation Error}
We observe the effect of translation errors by shifting a 500 x 500 detection left and right past a 500 x 500 square object centred within a 2000 x 2000 image.
This is tested both using BBoxes, and PBoxes with spherical Gaussian corners of varying reported variance (BBoxes equivalent to reported variance of zero).
The results from this test are shown in Figure~\ref{fig:eval_traits:translation}.

Here, we see that PDQ strongly punishes any deviation from the ground-truth for BBoxes with no spatial uncertainty.
In some cases PDQ drops close to zero after only a 10\% shift.
This is in strong comparison to mAP and moLRP which, while decreasing, does so at a far slower rate despite high confidence being supplied to incorrectly labelled pixels.
As a shift of 10\% is quite large for a 500 x 500 square, PDQ does not provide such leniency in its scoring until variance is 1000, at which point it closely follows the results of mAP and moLRP.
We see that as uncertainty increases, PDQ provides increased leniency, however, the highest score attainable drops reinforcing the idea that PDQ requires accurate detections with accurate spatial probabilities as stated within the main paper.

\begin{figure}[t]
    \centering
    \includegraphics[width=0.7\linewidth]{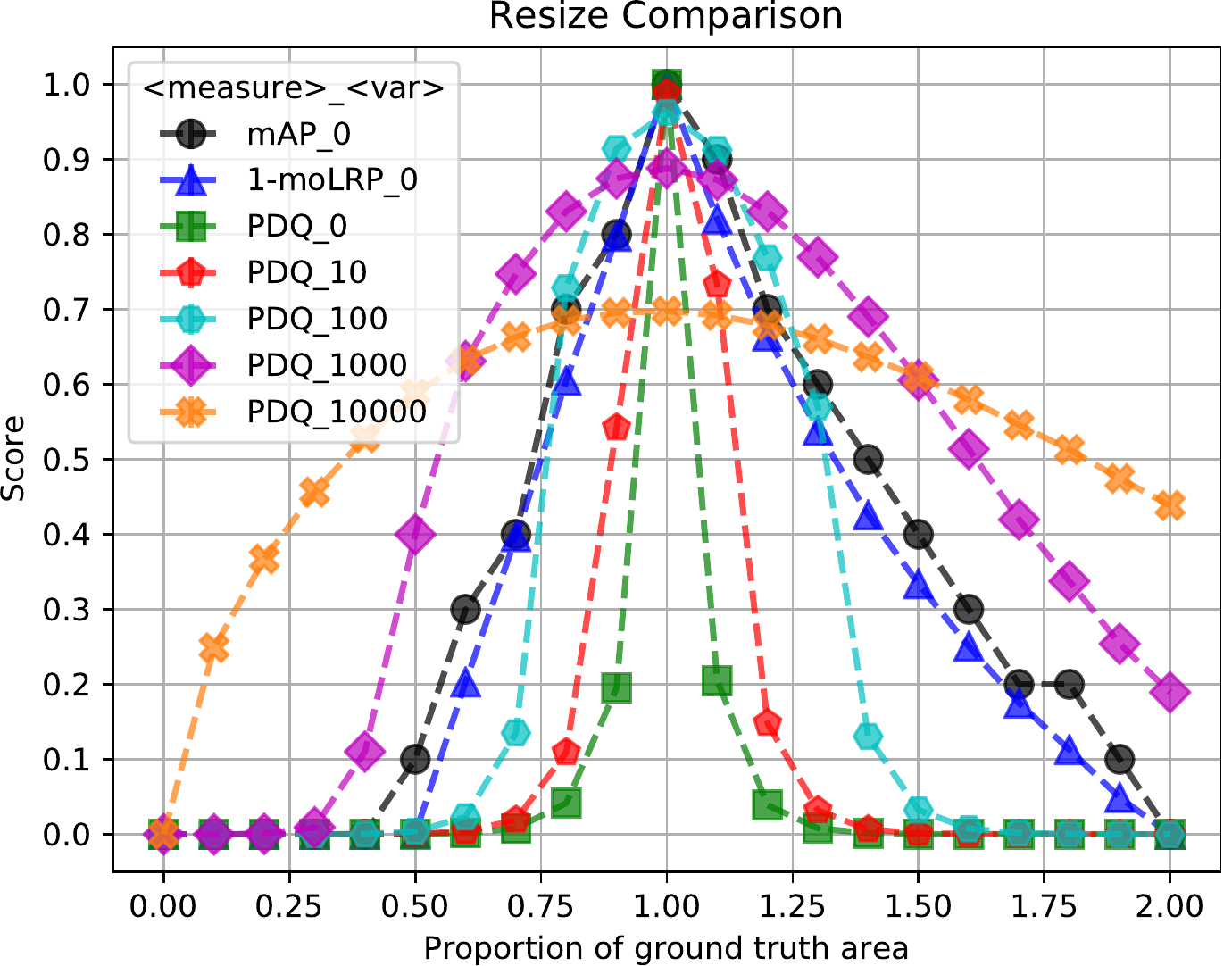}
    \caption{Evaluation of the effect of scaling on mAP, moLRP and PDQ scores. X-axis shows the proportional size of the detection to the ground-truth object. Variance (\textit{var}) refers to the variance of corner Gaussians of PBox detections. BBox is used when \textit{var} is zero. We see mAP and moLRP are lenient to detections with no uncertainty compared to PDQ and that PDQ is more lenient the more uncertain the detector is.}
    \label{fig:eval_traits:resize}
\end{figure}

\paragraph{Scaling Error}
Using the same experimental setup as the translation tests, rather than translating detections, we keep detections centred around the square object and adjust the corner locations such that the area of the square generated by them is proportionally bigger or smaller than the original object.
The results from this are shown in Figure~\ref{fig:eval_traits:resize}.

This reinforces the findings of the translation tests, showing how PDQ strongly punishes over-confidence or under-confidence in spatial uncertainty.
When there is greater deviation in box size, PDQ is more lenient when the uncertainty is higher.
We do not see this same response from mAP and moLRP which treat standard BBoxes with high confidence in a similar manner to PDQ on PBoxes with variance of 100.
We see from both this and the translation test that PDQ rewards boxes with high predicted variance  when the actual variance of the box is high.
This reinforces the finding of the main paper which states that PDQ requires accurate estimates of spatial uncertainty.

\subsection*{B.3. Label Quality}
\begin{figure}[t]
    \centering
    \includegraphics[width=0.8\linewidth]{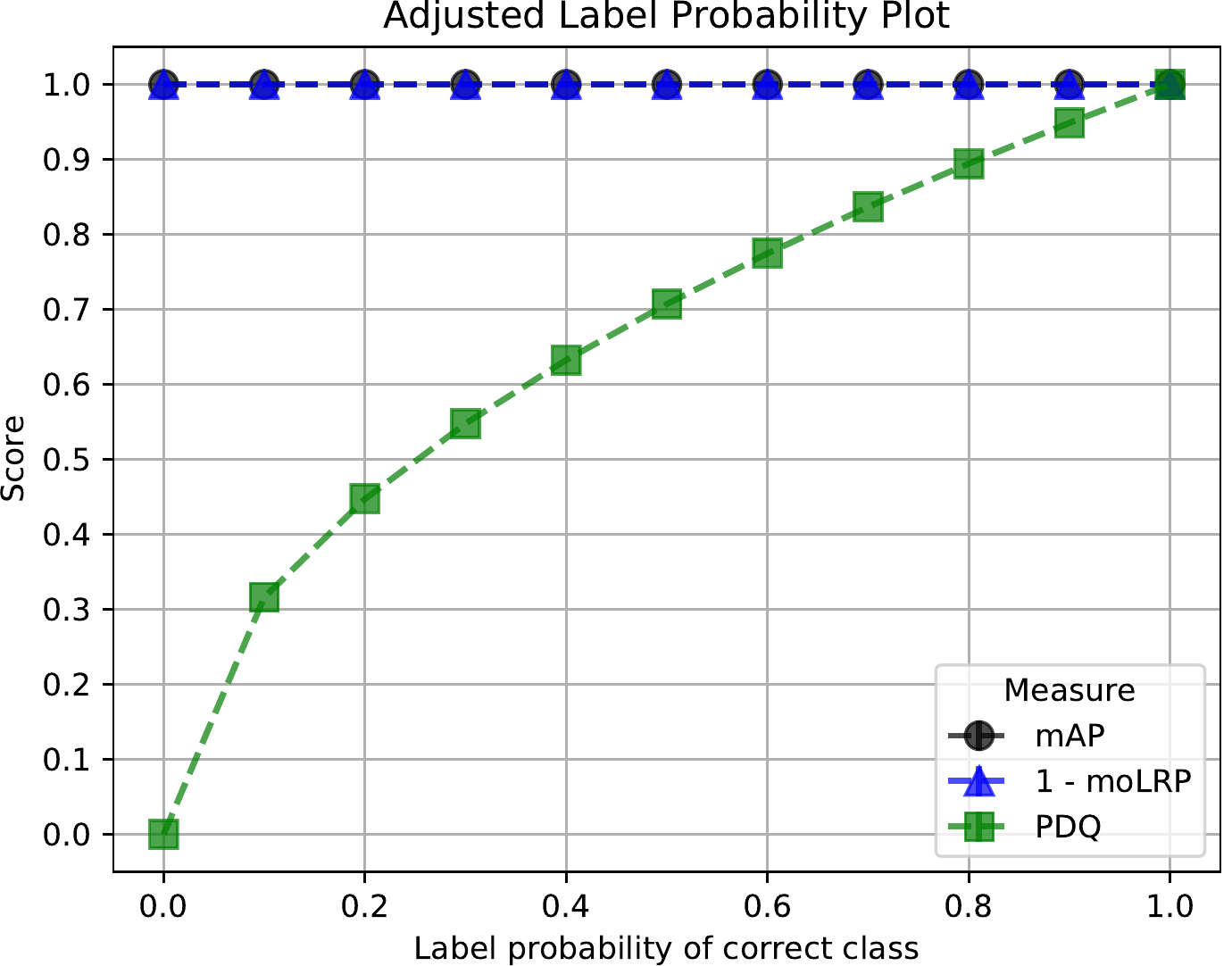}
    \caption{Effects of adjusting label confidences on mAP, moLRP, and PDQ when label probability for the correct class is adjusted using simulated detections on the COCO 2017 validation dataset. We see that existing measures are unaffected as long as the correct class is the class with highest probability in the label distribution. PDQ by comparison decreases with the label probability.}
    \label{fig:sup:reinforce:label}
\end{figure}

As demonstrated in the main paper, PDQ explicitly measures label quality, unlike existing measures.
We performed an additional test on the COCO 2017 validation data\cite{lin_microsoft_2014} using simulated detectors beyond that done in Section 6 of the main paper.
In this test, we set the label confidence for the correct class of each simulated detection to a given value and evenly distribute the remaining confidence between all possible other classes.
The results from this experiment when using perfectly aligned BBox simulated detections are shown in Figure~\ref{fig:sup:reinforce:label}.
This reinforces what had been seen previously, that existing measures are not explicitly effected by label probability, except when the maximum label confidence does not belong to the correct class.

\subsection*{B.4. Missed Ground-truth Objects}
\begin{figure}[t]
    \centering
    \includegraphics[width=0.7\linewidth]{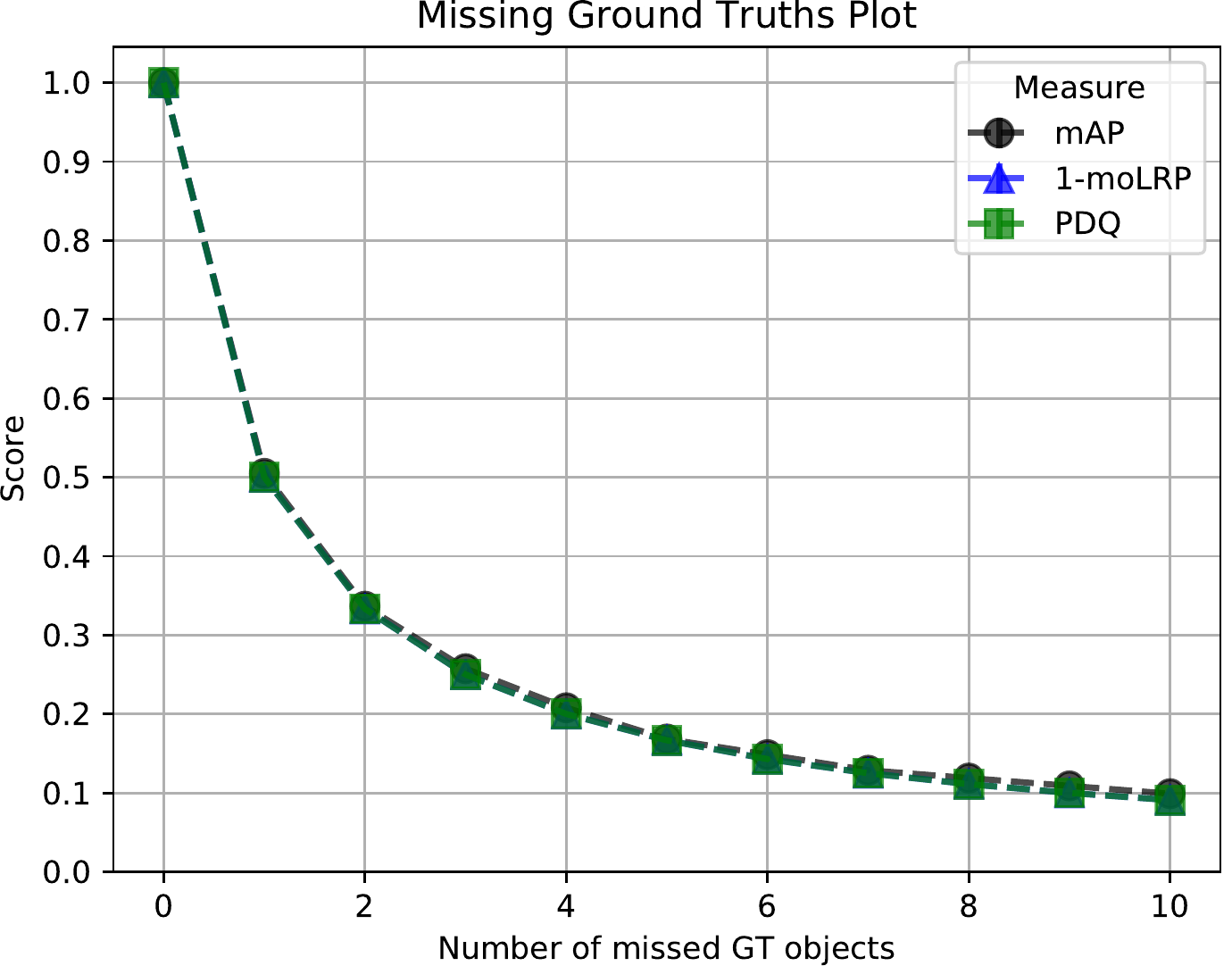}
    \caption{Evaluation of the effect of missing ground-truth objects on evaluation scores in simplified scenario. We observe that all measures respond the same to missed ground-truth objects.}
    \label{fig:eval_traits:miss_gt}
\end{figure}

\begin{figure}[t]
    \centering
    \includegraphics[width=0.8\linewidth]{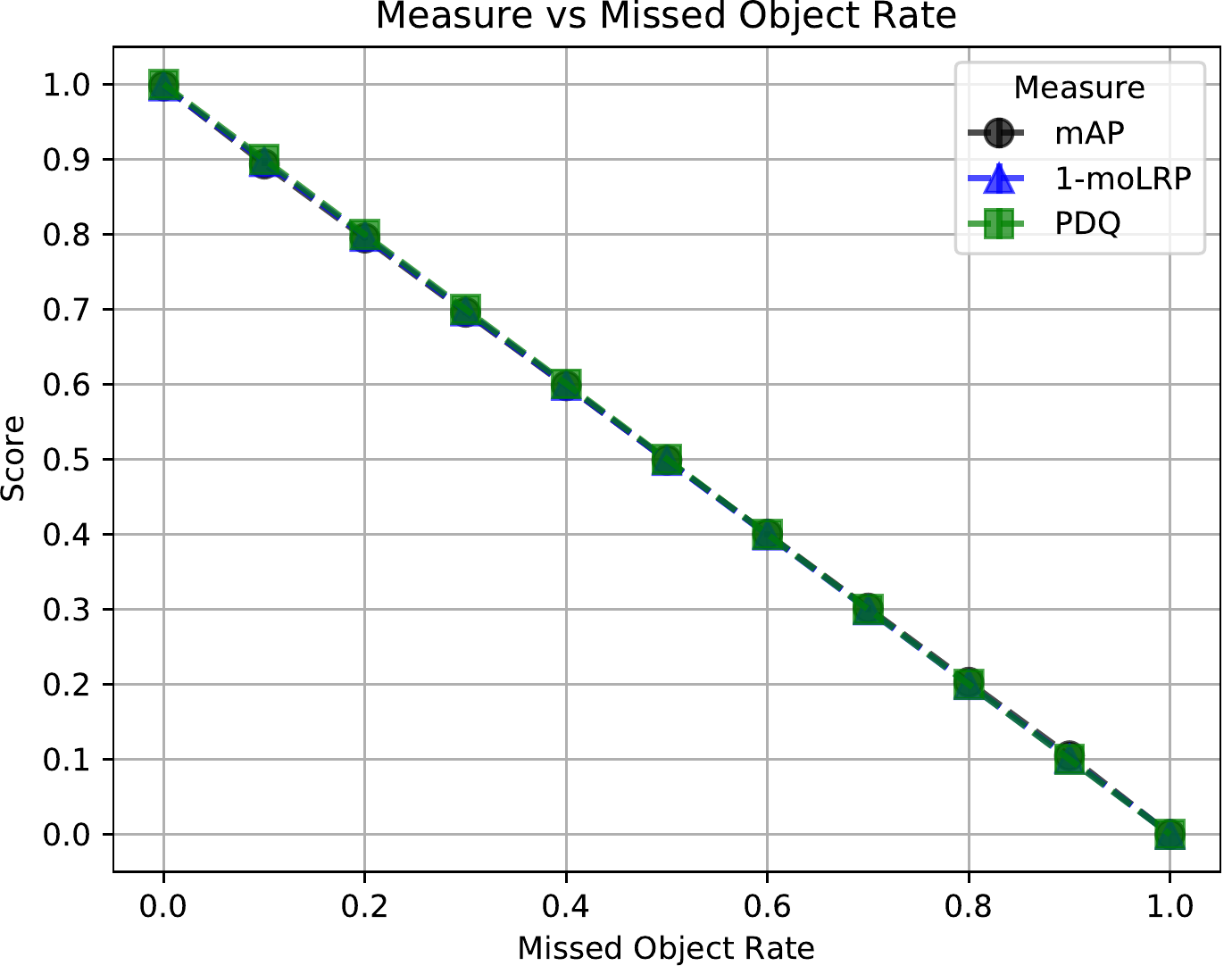}
    \caption{Evaluation of the effect of missing an increased proportion of ground-truth objects on COCO 2017 validation dataset images. We see the response from all measures is the same.}
    \label{fig:sup:reinforce:mor}
\end{figure}

We provide the results of two experiments that show that PDQ and existing measures perform the same when ground-truth objects are missed.
The first experiment is a simplified scenario where we add an increasing number of small 2 x 2 square objects around the edge of a single image with one large ground-truth object within it.
In this image, only the large ground-truth object is ever detected and the detection is spatially and semantically perfect.
Results for mAP, moLRP, and PDQ for this scenario are visualised in Figure~\ref{fig:eval_traits:miss_gt}.
The second experiment is performed on the COCO 2017 validation data using simulated detectors as done previously.
Here we define a missed object rate for all detectors which dictates the probability that a detection is generated for the given ground-truth object.
This was done for perfectly spatially aligned BBox detections and results can be seen in Figure~\ref{fig:sup:reinforce:mor}.

Both experiments show that, despite their other differences, mAP, moLRP, and PDQ respond the same to missed ground-truth objects (FNs).

\subsection*{B.5. False Detections}
\begin{figure}[t]
    \centering
    \includegraphics[width=0.7\linewidth]{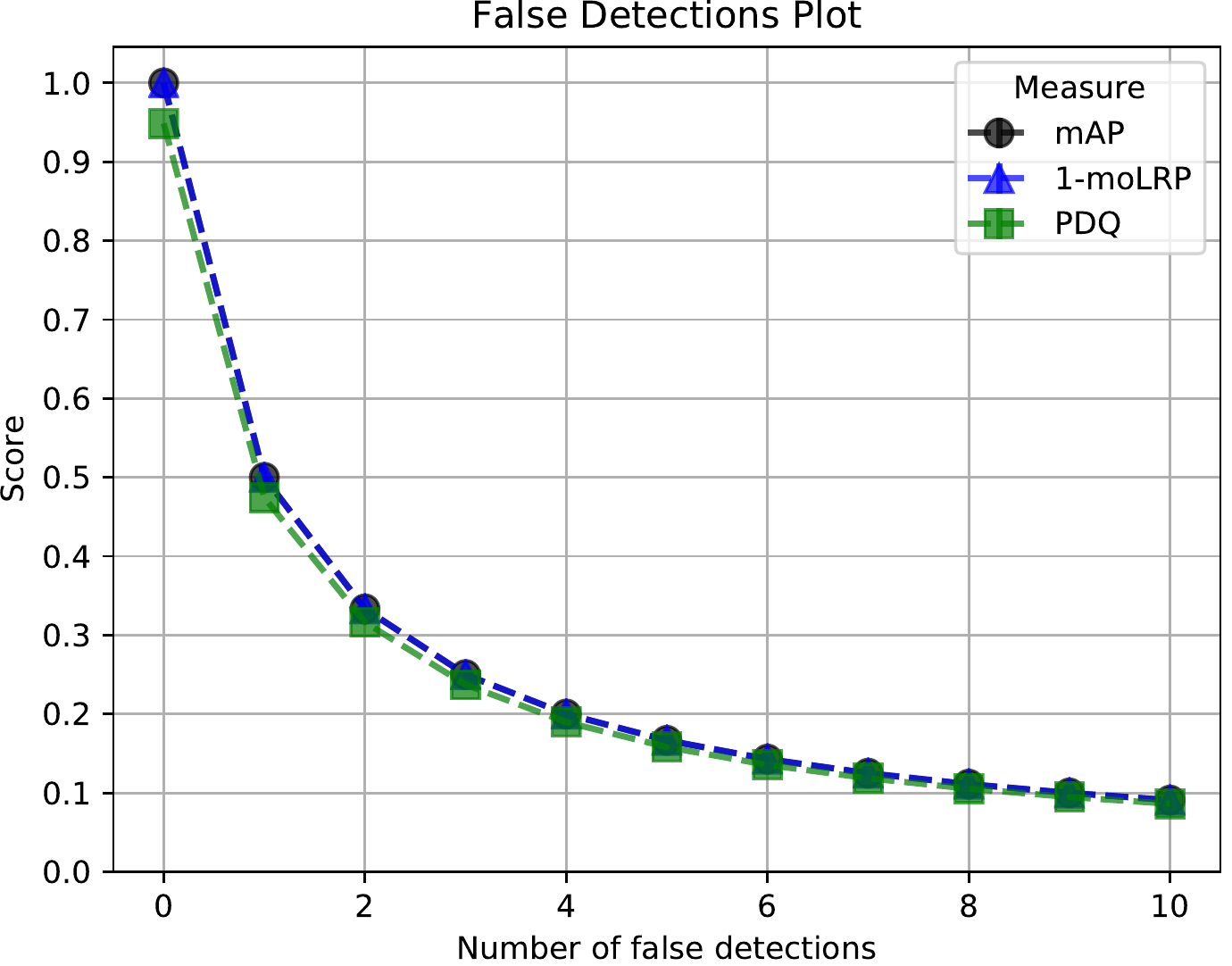}
    \caption{Evaluation of the effect of false detections on evaluation scores. We observe that generally, all measures respond the same to false detections.}
    \label{fig:eval_traits:duplication}
\end{figure}

We provide the results of a simplified scenario to show that, excluding edge cases that will be discussed in Section~\ref{sec:fp_obs}, mAP, moLRP, and PDQ respond almost the same to false positive detections.
To demonstrate this, we test a scenario where a single object in a single image is provided with a single perfectly spatially aligned detection and an increasing number of small 2 x 2 detections around the edge of the image.
The correct detection always has a label probability of 0.9 and all subsequent detections have a label probability of 1.0 so as to avoid edge cases for mAP explained and discussed in Section~\ref{sec:fp_obs}.
We plot the resultant mAP, moLRP, and PDQ scores in Figure~\ref{fig:eval_traits:duplication}.

Here we again observe consistency between the mAP, moLRP, and PDQ responses to false detections despite their differences in formulation.
Variations between PDQ and the other measures are caused by the lower label confidence for the correct detection which is known to effect PDQ.
While the responses here are almost identical, we have identified situations wherein mAP and moLRP obscure FP detections and lessen their impact.
\section*{C. Traditional Measures Obscuring False Positives.}\label{sec:fp_obs}

In the main paper, we describe how mAP and moLRP are able to obscure the impact of FPs present in the detections presented for evaluation.
To support these statements, we produce some simplified scenarios designed to demonstrate unintuitive outputs from mAP and moLRP when given FP detections.
Whilst not representative of how these measures are meant to act, they show unusual behavior for testing deployed detectors that PDQ does not share.
We do this through multiple test scenarios.

\subsection*{C.1. Duplicate 100\% Confident Detections}\label{sec:fp_obs:s1}
\begin{figure}[t]
    \centering
    \includegraphics[width=0.8\linewidth]{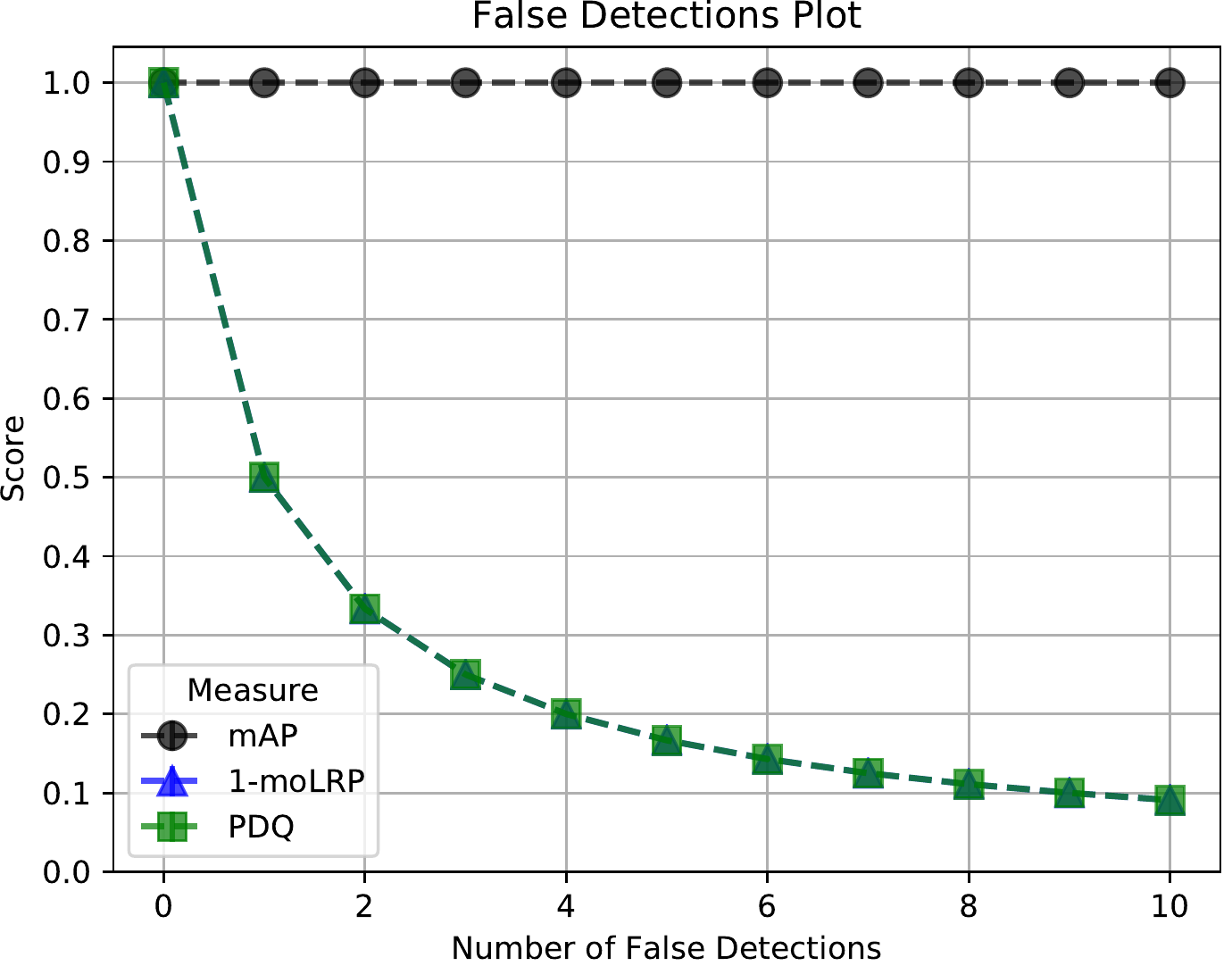}
    \caption{Duplications test results showing mAP, moLRP, and PDQ values when perfect duplicate FP detections are added in a one-object scenario. The TP detection is evaluated before the FPs, causing subsequent FPs to be ignored by mAP. PDQ and moLRP respond as expected, penalising FPs.}
    \label{fig:sup:edge:toy}
\end{figure}

In the first scenario, we consider detecting a single object in a single image where there is an increasing number of perfectly-aligned, 100\% confidence detections of that single object.
Results of this scenario are shown in Figure~\ref{fig:sup:edge:toy}.
We observed that PDQ and moLRP penalised the additional FP detections, whereas mAP gave 100\% accuracy at all times.

This edge case breaks mAP due to how the PR curve for this scenario is generated and utilised.
As is explained later in Section~\ref{sec:mAP}, the PR curve used for mAP uses the maximum precision at each level of recall to provide a smooth PR curve.
However, through this approach, it is assumed that as detections are added to the analysis, the result will be continually increasing recall.
Once the recall becomes perfect, or reaches some maximum value, any further false detections are ignored.
Here, as all detections have 100\% confidence and perfectly overlap the ground truth, the first detection is treated as the TP and all others are ignored.
The same effect would occur regardless of whether detections are perfect duplicates or located randomly within the image, as long as the TP is ordered first in confidence order (or in input order in the case of ties, see section \ref{sec:mAP}). 
This is why we attain the result for mAP shown in Figure~\ref{fig:sup:edge:toy}.

This is not a new problem with mAP, and such behaviour caused by relative ranking has been outlined in past works~\cite{redmon_yolov3:_2018}.
In comparison to this, moLRP and PDQ respond as expected to an increasing number of FP detections.
This is because both explicitly measure the number of false positives or the false positive rate from the detector output.
While robust to this first scenario, our second scenario shows that moLRP can also respond to false positive detections in the same unintuitive manner as mAP.

\subsection*{C.2. False Detections with Lower Confidence}\label{sec:fp_obs:s2}
\begin{figure}
    \centering
    \includegraphics[width=0.8\linewidth]{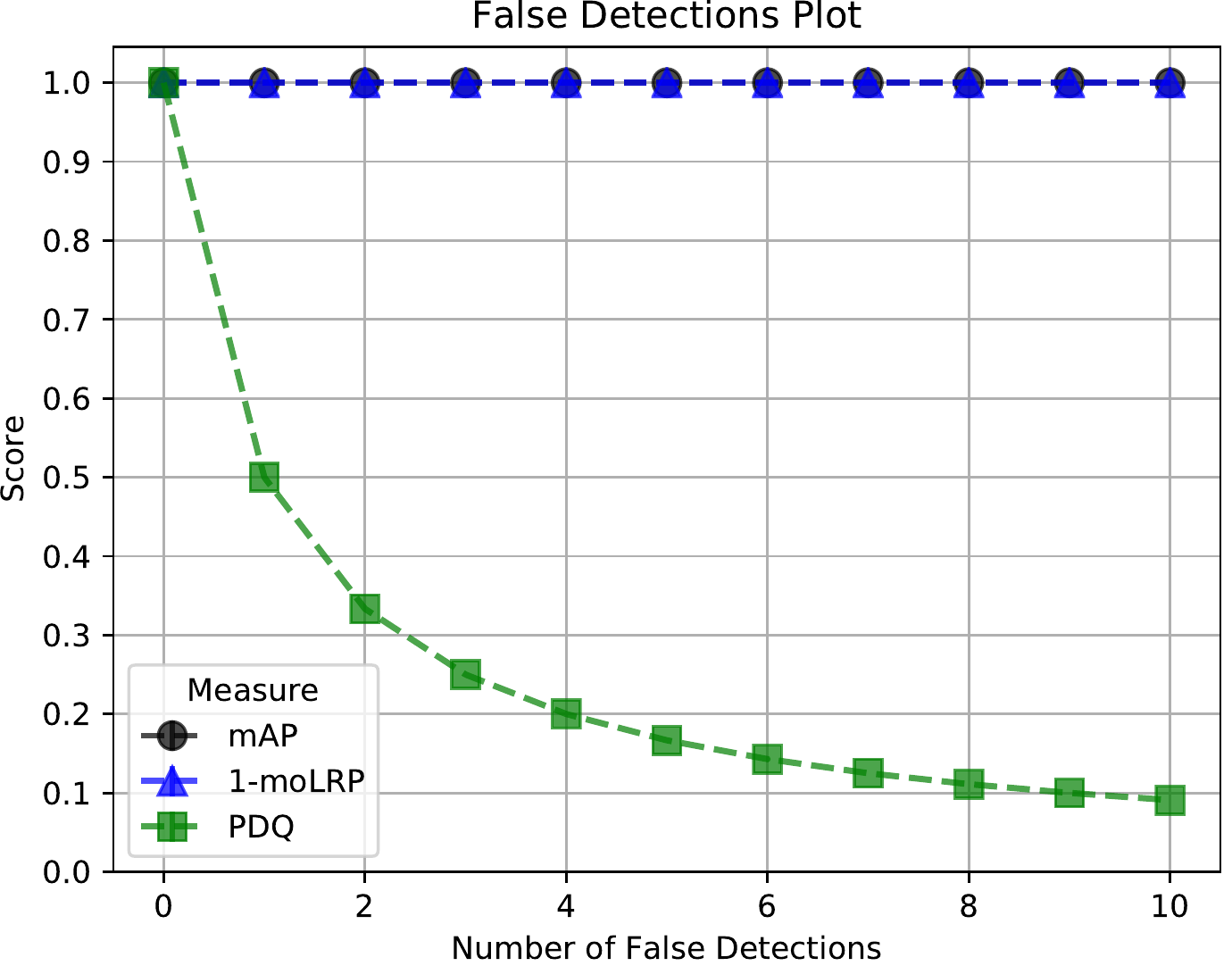}
    \caption{False detection test where all FP detections have slightly lower label confidence than the TP detection (90\% Vs 100\%). Both mAP and moLRP are shown to treat this as perfect detection output.}
    \label{fig:sup:edge:toy_harsh}
\end{figure}

Here, we consider a single image with a single object which is detected by a BBox detection of perfect spatial and semantic quality.
In addition to this, we introduce an increasing number of small false detections with label confidence 90\% around the border of the image.
The results from this scenario are shown in Figure~\ref{fig:sup:edge:toy_harsh}.

We observe in this scenario that mAP and moLRP both consider the results as perfect, regardless of the number of FPs, while PDQ penalises the increasing number of FP detections.
This mAP result comes from the same relative ranking issues as outlined in the previous scenario (Section \ref{sec:fp_obs:s1}).
The moLRP result, on the other hand, has changed due to the optimal thresholding done as part of the algorithm~\cite{oksuz2018localization}.
The moLRP score is designed to show the best possible performance of the detector if the best label confidence threshold for each class is chosen.
Choosing an ideal threshold above 0.9, the performance of the detector becomes perfect, despite the high-confidence false positive detections.
This trait of moLRP is beneficial for testing the ideal performance of a detector and for tuning a detector's final output.
However, as stated in the main paper this is not beneficial for testing systems to be applied in real-world applications, which cannot choose the optimal threshold on-the-fly during operation.
In contrast, PDQ does no such filtering and does not obscure false positive detections.

\subsection*{C.3. Duplicate Detections on COCO Data}\label{sec:fp_obs:s3}
\begin{figure}[t]
    \centering
    \includegraphics[width=0.8\linewidth]{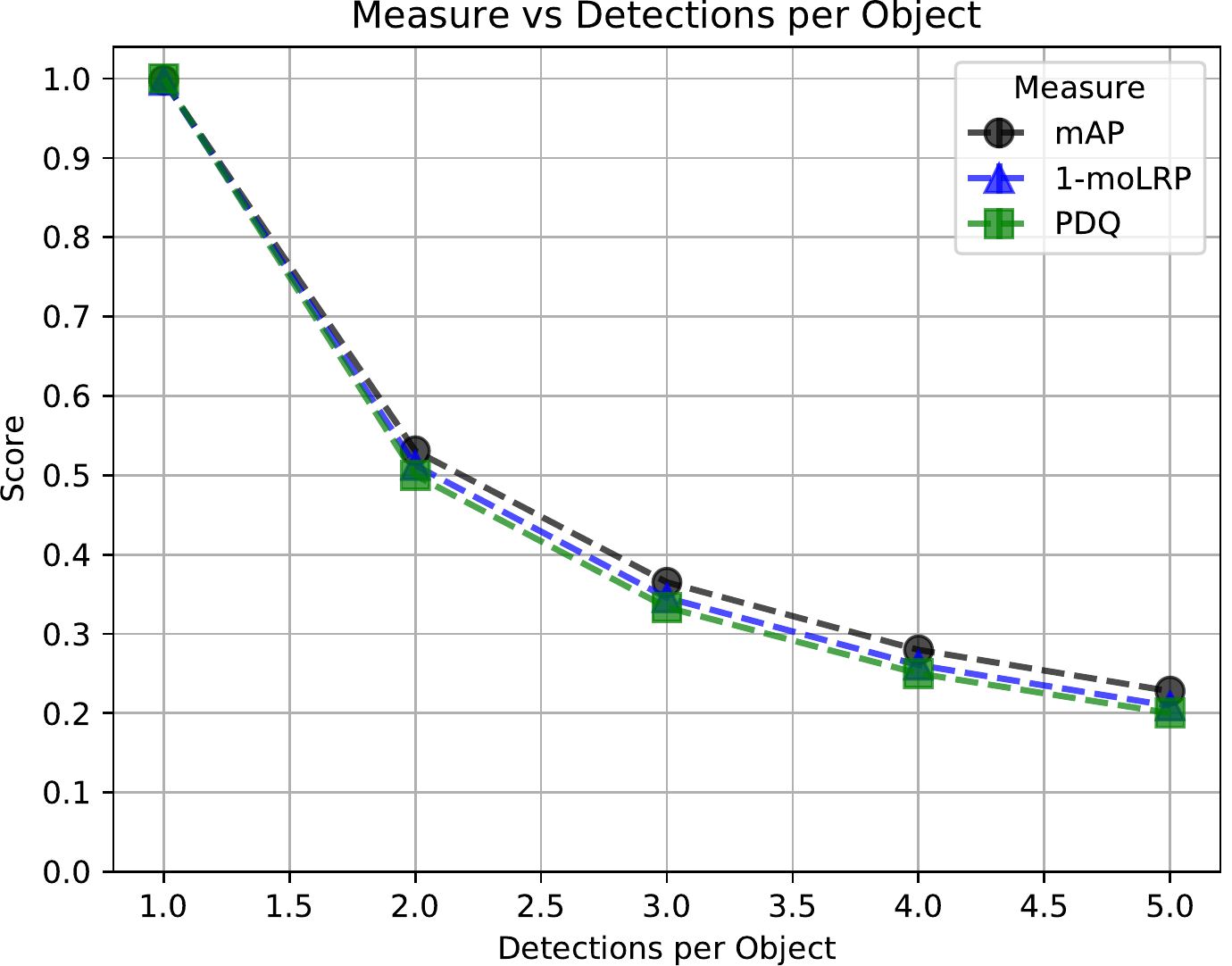}
    \caption{Test results on COCO 2017 validation data comparing scores when a number of perfectly aligned duplicate FP detections are added. Each duplicate FP is ordered directly after their corresponding TP detection. Due to smoothing of the PR curve, calculated precision becomes higher than expected for some classes at different levels of recall, causing mAP to be higher than expected. Other measures remain relatively unaffected.}
    \label{fig:sup:edge:coco}
\end{figure}

Scenario 3 extends scenario 1 (Section \ref{sec:fp_obs:s1}) from a single image to examine duplicate detections on the COCO 2017 validation data~\cite{lin_microsoft_2014}.
Again, every detection provided 100\% probability of being the correct class and was perfectly spatially aligned.
The detections are ordered such that all detections for a given object occur before the detections of the following object.
For example, if the number of duplicates is three, the order of detections would be three detections of object A followed by three detections of object B and so on. 
See Section \ref{sec:mAP} for why ordering is important.
It is expected that for such an experiment, the result for all evaluation measures would be reciprocal in nature (i.e. when there are 2 detections per object the score will be 1/2).
However, this is not exactly what we observed by our results as shown in Figure~\ref{fig:sup:edge:coco}.
What we see from this figure, is that the mAP provides scores slightly higher than expected, whereas PDQ and moLRP measures more closely follow the expected outcomes from such an experiment.

Again, this issue with mAP is caused by the smoothing of the PR curve outlined in Section~\ref{sec:mAP} and the ordering of our detections.
As described in Section \ref{sec:mAP}, mAP takes the maximum precision at each of its 101 sample recall values.
Additional FPs decrease precision, but don't affect the recall, and so are ignored.
As a simplified example, if two detections are given for every object, the recorded precision after 3 objects have been correctly detected is not 0.5 but rather 0.6 as three TPs have been evaluated to only two FPs, despite three FPs being present at this level of recall.
This can cause small discrepancies to occur and is the reason for mAP's unusual performance.
As we see in the following scenario, this is a problem which increases in severity with small datasets.

\subsection*{C.4. Duplicate Detections on Subset of COCO Data}\label{sec:fp_obs:s4}
\begin{figure}
    \centering
    \includegraphics[width=0.8\linewidth]{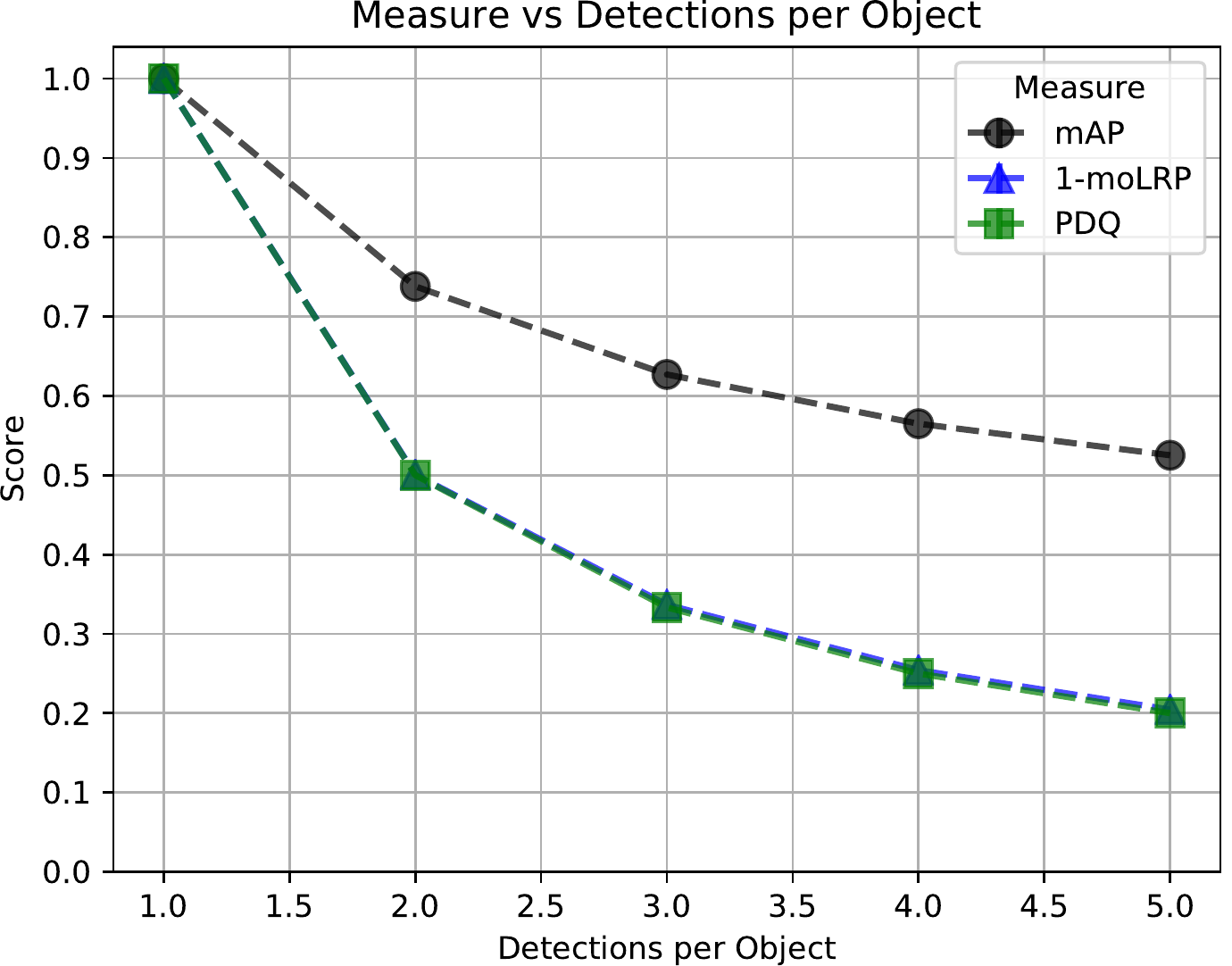}
    \caption{Duplication test results such as done for Figure~\ref{fig:sup:edge:coco} on subset of 100 images from COCO 2017 validation data. This shows heightened mAP scores from those shown Figure~\ref{fig:sup:edge:coco} demonstrating increased unintuitive behaviour from mAP as the dataset gets smaller.}
    \label{fig:sup:edge:coco_heightened}
\end{figure}

In the fourth scenario, we increase the severity of the mAP error found in the previous scenario (Section \ref{sec:fp_obs:s3}).
We do this by testing on a subset of the full 5,000 COCO images previously used,  evaluating on only the first 100 images.
We show these results in Figure~\ref{fig:sup:edge:coco_heightened}.

Here we see that the mAP scores are far higher at than expected for each level of detections per object, an exaggeration of the effect in Section \ref{sec:fp_obs:s3}.
This occurs because the smaller number of ground truth instances results in fewer possible measurable recall values.
As precision is recorded at 101 set levels of recall, and (as established in Section \ref{sec:fp_obs:s3}) FPs are obscured until a new measured level of recall is reached, the FPs remain obscured for more recorded levels of recall.
Correspondingly, there are fewer total detections at each recorded level of recall, making the number of obscured FPs relatively more significant.
This means that more of the recorded maximum precision values are higher, leading to a higher mAP score.

This can ultimately result in the extreme case discussed in Section \ref{sec:fp_obs:s1}.
We observe then that the issues caused by the obfuscation of FPs under mAP increases as the number of samples tested gets smaller.
Again, we note that both moLRP and PDQ do not suffer from this issue, as they explicitly measure FPs.

\subsection*{C.5. Duplicate Detections with Lower Confidence on COCO Data}\label{sec:fp_obs:s5}
\begin{figure}
    \centering
    \includegraphics[width=0.8\linewidth]{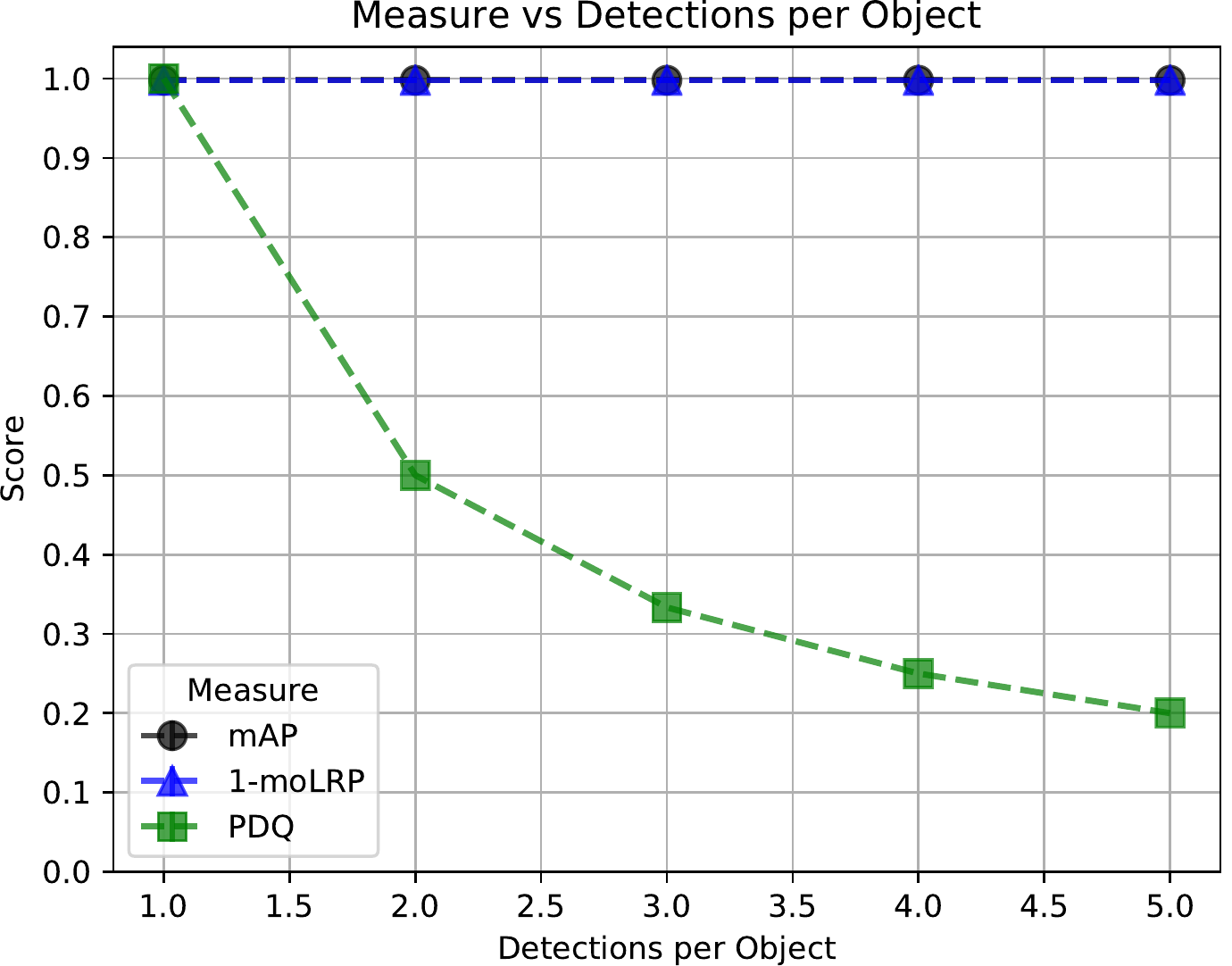}
    \caption{Duplication test results on COCO 2017 validation data where all FP duplicate detections have lower label confidence than the TP detection (90\% Vs 100\%). Unlike PDQ, both mAP and moLRP are shown to treat this as perfect detection output.}
    \label{fig:sup:edge:coco_harsh}
\end{figure}

Reinforcing our findings in Sections \ref{sec:fp_obs:s3} and \ref{sec:fp_obs:s4}, we show again that moLRP, while sometimes avoiding pitfalls present in mAP, can still obscure false positive detections through optimal thresholding.
In this scenario, we ensure that only the first detection has label confidence of 100\% and all subsequent duplicate detections have label confidence of 90\%.
The results of this test are shown in Figure~\ref{fig:sup:edge:coco_harsh}.
As expected, PDQ continues to treat the false positives as significant whilst mAP and moLRP both consider the detection output as perfect.

\subsection*{C.6. Summary}\label{sec:fp_obs:summary}
In summary, we have demonstrated extreme scenarios showing that both mAP and moLRP can obscure false positive detections under different conditions leading to unintuitive results.
These issues result from the assumptions made when generating and using PR curves for mAP and optimal thresholding for moLRP.
As stated in the main paper, this unintuitive nature is inappropriate behavior for evaluating detectors meant for real-world deployment.
We show that PDQ is unaffected by such scenarios, reinforcing the findings of the main paper.
\begin{figure*}[t]
    \centering
    \includegraphics[width=0.8\linewidth]{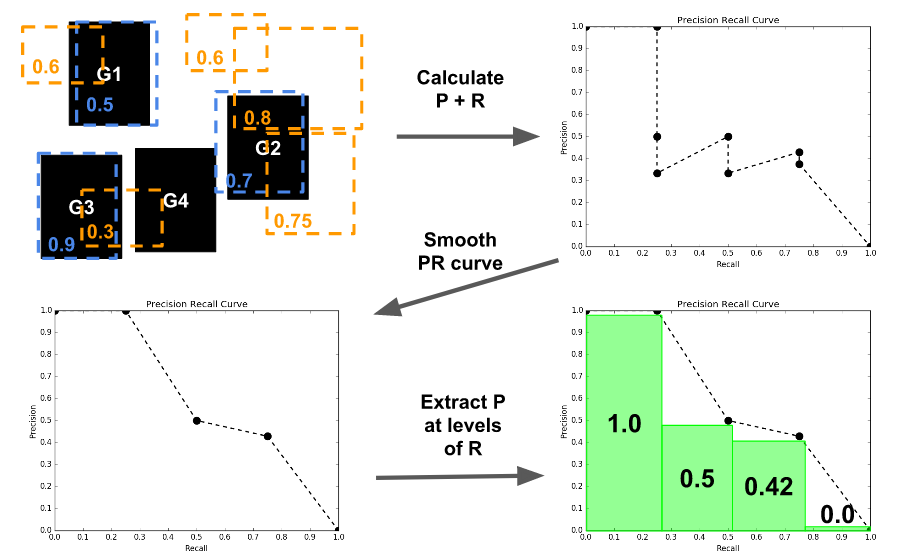}
    \caption{Process for extracting precision values from a PR curve for a given object class at a given threshold. The top-left shows the example scenario with ground-truth objects shown as black boxes, true-positive detections shown as light blue BBoxes, and false-positive detections shown as orange BBoxes. Numbers within the boxes represent label confidence. Top-right figure shows PR curve generated as each detection is added in order of decreasing label confidence. Bottom-left figure shows the effect of smoothing the PR curve by only taking the maximum precision values. Bottom-right shows the precision values extracted for a given range of recall values examined. Note that 101 samples are made across different levels of recall. Best viewed in colour.}
    \label{fig:mAP}
\end{figure*}
\section*{D. Definition of mAP}\label{sec:mAP}
For the sake of completeness and to aid in understanding the behaviour shown in Section~\ref{sec:fp_obs}, here we define mean average precision (mAP) as used by the COCO detection challenge~\cite{lin_microsoft_2014}.
Each detection provides a bounding box (BBox) detection location ($\cB^f_j$) and a confidence score for its predicted class $s^f_j$.
For each detection in the $f$-th frame of a given class, mAP assigns detections to ground-truth objects of that same class.
Each detection is defined as as either a true positive (TP) if it is assigned to a ground-truth object, or a false positive (FP) if it is not.
Detections for each class are ranked by confidence score and assigned to ground-truth objects in a greedy fashion if an intersection over union (IoU) threshold $\tau$ is reached.
IoU is calculated as follows

\begin{equation}
    IoU(\hat{\cB}^f_i, \cB^f_j) = \frac{area(\hat{\cB}^f_i \cap \cB^f_j)}{area(\hat{\cB}^f_i \cup \cB^f_j)} ,
\end{equation}

\noindent where $\hat{\cB}^f_i \cap \cB^f_j$ is the intersection of the ground-truth and detection bounding boxes and $\hat{\cB}^f_i \cup \cB^f_j$ is their union.
The assignment process is summarized by Algorithm 1 and results in an identity vector $\textbf{z}$ which describes for each detection, whether it is a TP or FP with values of 1 or 0 respectively.

\begin{algorithm}[t]
\SetAlgoLined\SetArgSty{}
\KwData{a dataset of $f=1\dots N_F$ frames with detections $\cD^f = \{ \cB^f_j, s^f_j \}_{j=1}^{N^f_D}$ and ground truths $\cG^f = \{ \hat{\cB}^f_i\}_{i=1}^{N^f_G}$ for each frame for a given class $\hat{c}$}
Let $\cU$ be the set of unmatched objects \\
\ForAll{frames in the dataset}{
    order detections by descending order of $s^f_j$ \\
   \ForAll{detections in frame}{
        $\cG^f_* = \argmax_{\cG^f_i} IoU(\cG^f_i, \cD^f_j)$
        \uIf{$IoU (\cG^f_*, \cD^f_j) > \tau$ and $\cG^f_* \in \cU$}{
            $z^f_j = 1$ \\
            $\cU = \cU - \cG^f_*$
        }
        
    }
}
Return $\textbf{z} = [z^1_1, z^1_2, \dots, z^{N_F}_{N^{N_F}_D} ]$
\caption{mAP Detection Assignment}
\end{algorithm}

After the assignment process is conducted for all images, a precision-recall (PR) curve is computed from the ranked outputs of the given class.
Precision and recall are calculated for each detection as it is ``introduced'' to the evaluation set in order of highest class confidence (and then in submission order in the event of confidence ties).
Precision is defined as the proportion of detections evaluated that were true positives, and recall is defined as the proportion of ground-truth objects successfully detected.
After generating the PR curve for the given class, the maximum precision is recorded for 101 levels of recall uniformly spaced between zero and one.
The maximum precision is used to avoid ``wiggles'' in the PR curve, resulting in a smoothed PR curve.
If no precision has been measured for a given level of recall, the precision at the next highest measured level of recall is recorded.
Maximum precision at recall values above the highest reached are 0, to handle false negatives (FNs).
This process on a simple scenario is outlined visually in Figure~\ref{fig:mAP}.
This is process repeated for every evaluated class and at multiple values of $\tau$.
The average of all recorded precision values across all IoU thresholds, classes, and recall levels, provides the final mAP score.

\end{document}